%% file: main.tex
\documentclass[10pt,twocolumn,letterpaper]{article}
\PassOptionsToPackage{table}{xcolor}
\usepackage[pagenumbers]{cvpr}
\input{preamble}
\title{DisasterVLG: Vision-Language Grounding and Evidence-Based Reasoning for Disaster Scene Understanding}
\author{Ali Diba$^{1}$, Aisha Al-Mohannadi$^{1,2}$, Yin Yang$^{2}$, Muhammad Imran$^{1}$, Ferda Ofli$^{1}$\\
$^{1}$Qatar Computing Research Institute, Hamad Bin Khalifa University, Doha, Qatar\\
$^{2}$College of Science \& Engineering, Hamad Bin Khalifa University, Doha, Qatar\\
{\tt\small \{adiba,aalmohannadi,yyang,mimran,fofli\}@hbku.edu.qa}}

\begin{document}
\maketitle
\input{sections/abstract}

\section{Introduction}
\input{sections/intro}

\section{Related Work}
\input{sections/related}
\section{DisasterVLG Dataset}
\input{sections/dataset}

\section{DisasterVLG Tasks and Models}
\input{sections/tasks_models}
\section{Experiments}
\input{sections/experiments}

\section{Conclusion}

\input{sections/conclusion}

{
\small
\bibliographystyle{ieeenat_fullname}
\bibliography{main}
}

\clearpage
\begin{center}
\textbf{Supplementary Material}
\end{center}
\appendix
\input{sections/appendices}

\end{document}

%% file: preamble.tex
\usepackage[utf8]{inputenc}
\usepackage[T1]{fontenc}
\usepackage{nicefrac}
\usepackage{microtype}
\usepackage{multirow}
\usepackage{makecell}
\usepackage{enumitem}
\usepackage[most]{tcolorbox}
\usepackage{listings}
\definecolor{headerlavender}{RGB}{198,198,236}

\definecolor{metricblue}{RGB}{62,125,190}
\definecolor{gainblue}{RGB}{225,237,250}

\newcommand{\grad}[2]{\cellcolor{metricblue!#1}#2}
\newcommand{\gain}[1]{\cellcolor{gainblue}\textbf{#1}}

\definecolor{promptbg}{RGB}{245,247,252}
\definecolor{promptframe}{RGB}{120,140,200}

\lstdefinestyle{promptlisting}{
  basicstyle=\ttfamily\scriptsize,
  breaklines=true,
  breakatwhitespace=false,
  columns=fullflexible,
  keepspaces=true,
  showstringspaces=false,
  frame=none,
  xleftmargin=0pt,
  xrightmargin=0pt
}

\newtcblisting{promptbox}[1][]{
  enhanced,
  breakable,
  colback=promptbg,
  colframe=promptframe,
  boxrule=0.8pt,
  arc=2mm,
  left=1mm,
  right=1mm,
  top=1mm,
  bottom=1mm,
  fonttitle=\bfseries,
  coltitle=white,
  title=#1,
  listing only,
  listing engine=listings,
  listing options={style=promptlisting},
  attach boxed title to top left={xshift=0mm,yshift=0mm},
  boxed title style={
    colback=promptframe,
    colframe=promptframe,
    arc=2mm,
    boxrule=0pt
  }
}

\definecolor{cvprblue}{rgb}{0.21,0.49,0.74}
\usepackage[pagebackref,breaklinks,colorlinks,allcolors=cvprblue]{hyperref}
\def\confName{CVPR}
\def\confYear{2027}

%% file: sections/abstract.tex
\begin{abstract}
General-purpose vision-language models struggle with crisis imagery due to its inherent noise, ambiguity, and high variability across disaster events --- yet timely, accurate scene understanding is critical for humanitarian response. To address this, we introduce \textbf{DisasterVLG}, a large-scale multi-task dataset containing \textbf{1.24 million} disaster images from \textbf{142 countries} paired with \textbf{33 million} structured annotations spanning dense captioning with localization, incident assessment, relational scene descriptions, and crisis-grounded visual question answering (VQA) across
diverse hazard types and scene contexts worldwide. Unlike prior disaster benchmarks that focus narrowly on classification, DisasterVLG unifies recognition, grounded reasoning, and uncertainty-aware interpretation. A distinguishing feature is its \textbf{evidence-grounded} annotation design: each VQA instance includes supporting evidence text that links visual observations to answers, alongside positive and negative question pairs, multiple answer formats, and difficulty labels. Visual grounding layer links caption phrases to image regions and relationship statements to subject-object boxes, with explicit labels for visual supports and clues. To support rigorous evaluation, we construct a human-verified test set of \textbf{38K} images (\textbf{1.15M} annotations) spanning broad geographic regions and disaster types. Experiments with domain-adapted vision-language models demonstrate strong performance across disaster captioning, incident recognition, and grounded VQA, while also exposing the limitations of general-purpose models in crisis-specific reasoning. DisasterVLG establishes a scalable, reproducible benchmark for vision-language models that must reason reliably under real-world uncertainty. We will publicly release the dataset, trained models, and code.
\end{abstract}

%% file: sections/intro.tex
\label{sec:intro}

Visual evidence is critical to disaster response and humanitarian coordination, as it offers rapid insight into impacts that are often difficult to capture through textual reports alone~\cite{lagerstrom2016image,nguyen2017damage,alam2017image4act}. Images shared during crisis incidents, such as floods, earthquakes, can reveal damaged buildings, inundated roads, blocked infrastructure, displaced populations, and other critical conditions on the ground~\cite{alam2021humaid,alam2023medic}. As the volume of such imagery continues to grow across online social platforms, news media, and field reports, the ability to automatically interpret disaster images at scale has become increasingly important. Reliable visual understanding can support faster situational awareness, improve prioritization of response actions, and help humanitarian organizations make more informed decisions under time pressure and uncertainty~\cite{asif2021automatic,pennington2022near,imran2022ai,bashir2024efficient}.

Recent advances in vision-language models offer a promising path toward automated disaster image understanding, enabling systems to describe images, answer questions, and support flexible interaction with visual data~\cite{radford2021clip,instructblip,llava}. Yet crisis imagery remains challenging for general-purpose VLMs because such images are often noisy, ambiguous, occluded, and shaped by complex cues such as smoke, debris, and floodwater~\cite{alam2022robust,gupta2025disasterres}. In these settings, models may rely on superficial patterns rather than grounded visual evidence, producing hallucinated objects and overconfident but incorrect answers to operationally relevant questions~\cite{goyal2017making,gunjal2024detecting}. Such errors are especially consequential in humanitarian contexts, where visual misinterpretation can distort situational awareness and downstream decisions.
Existing disaster vision datasets, such as CrisisMMD~\cite{alam2018crisismmd}, Incidents1M~\cite{weber2023incidents1m}, xBD~\cite{gupta2019xbd}, have enabled progress in incident classification, damage assessment, and humanitarian information extraction. However, most remain designed around narrow recognition tasks or specific data modalities, offering limited support for evaluating broader visual understanding, grounded reasoning, and generalization across diverse disaster events.

\begin{figure*}[t]
    \centering
    \includegraphics[width=1\linewidth]{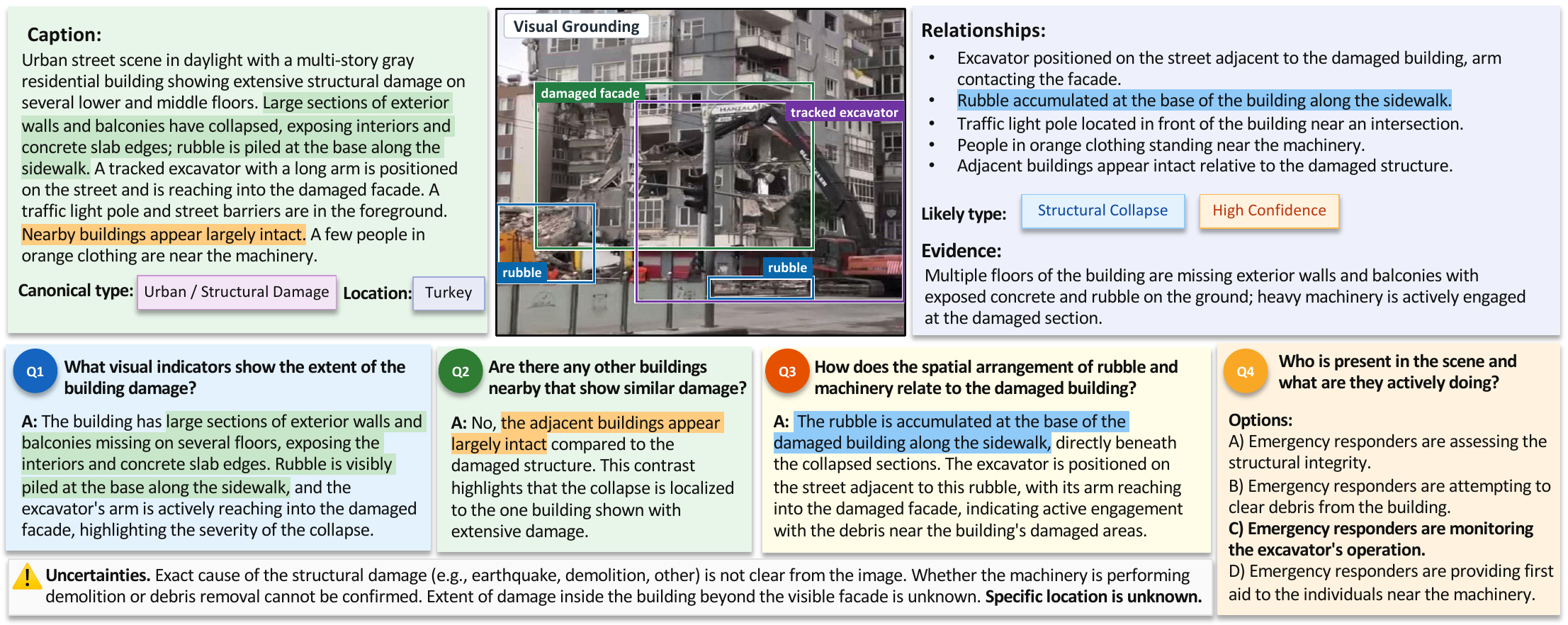}
    \caption{A DisasterVLG sample and its multi-layer annotations: caption, disaster type, evidence, spatial relationships, question-answer pairs, multiple-choice questions, and uncertainty statements.}
    \label{fig:intro}

\end{figure*}

To address this gap, we introduce \textbf{DisasterVLG}, a large-scale dataset for grounded vision-language learning in disaster scenarios. DisasterVLG contains approximately 1.2 million crisis-related images from 142 countries organized under a canonical taxonomy of 29 disaster/hazard categories, covering more than 12K incident types, such as wildfires, earthquakes, urban destruction, and floods. More importantly, DisasterVLG provides structured annotations for complementary tasks, including \textit{(i)} image captioning (1.2 million), \textit{(ii)} incident classification and reasoning (1.2 million), \textit{(iii)} grounded visual question answering (20.5 million), and \textit{(iv)} multiple-choice question answering (2.7 million). These annotations are generated through a controlled VLM-assisted pipeline that emphasizes observable visual evidence, uncertainty marking, and humanitarian information needs such as hazard recognition, damage indicators, spatial exposure, infrastructure conditions, response presence, and hallucination resistance. This design enables the same visual content to be studied from multiple perspectives, ranging from descriptive understanding to grounded and operationally meaningful crisis reasoning.

We extend the annotation with region-level visual grounding: visual clues from caption text are linked to bounding boxes, and relationship statements are linked to subject and object regions with supported, unclear, or contradicted labels. Phrases that cannot be localized are retained with a reason. This adds an explicit localization task to the existing captioning and reasoning tasks, enabling inspection of the spatial evidence associated with a textual claim.

To assess different levels of disaster image understanding, we evaluate DisasterVLG by training VLMs on the dataset and testing their performance across image captioning, VQA, and reasoning tasks. We fine-tune Qwen3.5 models on DisasterVLG and compare them against general-purpose open-source and proprietary VLMs using task-appropriate evaluation metrics. The results show that DisasterVLG fine-tuning improves disaster-specific visual understanding: the adapted 9B model outperforms all baselines, including GPT-5, on every generative and grounding task, with the largest gains on reference-overlap metrics for reasoning tasks. Gains are smaller on GES, which scores factual agreement rather than wording, and on multiple-choice QA, where adapted and strong general-purpose models perform similarly.

Overall, DisasterVLG provides a large-scale foundation for disaster-focused vision-language learning. To the best of our knowledge, it is the first multimodal disaster dataset to combine million-scale captions, tens of millions of VQA pairs, broad geographic coverage, and diverse disaster types within a unified benchmark. Models trained on DisasterVLG show substantial gains over general-purpose baselines, demonstrating the value of domain-specific, evidence-oriented supervision for reliable disaster image understanding.

%% file: sections/related.tex
\label{sec:related}
\newcommand{\cyes}{\textcolor{green!60!black}{\checkmark}}
\newcommand{\cno}{\textcolor{red!70!black}{$\times$}}
\newcommand{\cpartial}{\textcolor{orange!80!black}{$\circ$}}

\begin{table*}[t]
\centering
\caption{Comparison of disaster-oriented datasets by view, scale and task coverage.
}
\label{tab:dataset_comparison}
\small
\setlength{\tabcolsep}{4pt}
\resizebox{\textwidth}{!}{%
\begin{tabular}{l l r r c c c c c c c}
\toprule
\textbf{View} & \textbf{Dataset} & \textbf{Images} & \textbf{Ann.} & \textbf{Classification} & \textbf{Segmentation} & \textbf{Captioning} & \textbf{VQA} & \textbf{MCQ} & \textbf{Visual Grounding} & \textbf{Reasoning} \\
\midrule
\multirow{5}{*}{Aerial}
& Sen1Floods11~\cite{bonafilia2020sen1floods11} & 4,831 & - & \cno & \cyes & \cno & \cno & \cno & \cno & \cno \\
& Sen12Landslides~\cite{hohn2025spatio}         & 39K   & - & \cno & \cyes & \cno & \cno & \cno & \cno & \cno \\
& xBD~\cite{gupta2019xbd}                     & 22K   & - & \cno & \cyes & \cno & \cno & \cno & \cno & \cno \\
& FloodNet~\cite{rahnemoonfar2021floodnet}     & 2,348 & 7,345 & \cyes & \cyes & \cno & \cyes & \cno & \cno & \cno \\
& DisasterM3~\cite{wang2026disasterm3}        & 27K   & 123K  & \cyes & \cyes & \cyes & \cno & \cyes & \cno & \cyes \\
\midrule
\multirow{5}{*}{Ground}
& MEDIC~\cite{alam2023medic}                  & 72K   & -     & \cyes & \cno & \cno & \cno & \cno & \cno & \cno \\
& Incidents1M~\cite{weber2023incidents1m}      & 977K  & -     & \cyes & \cno & \cno & \cno & \cno & \cno & \cno \\
& CrisisMMD~\cite{alam2018crisismmd}          & 18K   & 16K   & \cyes & \cno & \cno & \cno & \cno & \cno & \cno \\
& DisasterVQA~\cite{almohannadi2026disastervqa} & 1,395 & 4,405 & \cno & \cno & \cno & \cyes & \cyes & \cno & \cno \\
& \textbf{DisasterVLG (Ours)} & \textbf{1.24M} & \textbf{33M} & \cyes & \cno & \cyes & \cyes & \cyes & \cyes & \cyes \\
\bottomrule
\end{tabular}%
}
\end{table*}

\paragraph{Disaster Image Analysis Datasets.}

Prior disaster image understanding broadly splits into aerial/satellite and ground-level analysis. Remote sensing (RS) datasets have targeted specific disaster types, such as floods (Sen1Floods11~\cite{bonafilia2020sen1floods11}) and landslides (Sen12Landslides~\cite{hohn2025spatio}), or specific tasks such as building damage assessment (xBD~\cite{gupta2019xbd}). More recent benchmarks like FloodNet~\cite{rahnemoonfar2021floodnet} and DisasterM3~\cite{wang2026disasterm3} have incorporated VQA for overhead visual reasoning, yet remain inherently limited by a bird's-eye perspective that misses proximal humanitarian signals. Ground-level datasets have similarly evolved from specialized damage assessment toward broader multimodal categorization: MEDIC~\cite{alam2023medic} and CrisisMMD~\cite{alam2018crisismmd} established foundations for humanitarian classification, while Incidents1M~\cite{weber2023incidents1m} provided scale for robust detection. However, these resources remain predominantly classification-centric. Nascent reasoning-oriented benchmarks like DisasterVQA~\cite{almohannadi2026disastervqa} show promise but lack the scale required to fine-tune modern VLMs (see Table~\ref{tab:dataset_comparison}). DisasterVLG bridges this gap by unifying large-scale incident detection with generative reasoning --- integrating captioning, incident labeling, and open-ended and multiple-choice VQA to enable grounded, reliable crisis reasoning.

\paragraph{Visual Question Answering and Grounded Reasoning.}
VQA has evolved from basic recognition toward compositional reasoning and visual grounding. Early benchmarks exposed a critical flaw: models could exploit answer priors without genuinely attending to images~\cite{antol2015vqa,goyal2017making,agrawal2018dont}, prompting a shift toward balanced distributions, polarity-aware evaluation, and evidence-based supervision~\cite{ramakrishnan2018overcoming,hudson2019gqa,han2020finding,radford2021learning,chen2023vqa,wang2026vgr}. These principles are particularly salient for disaster analysis, where negative evidence and uncertainty-aware reasoning are as critical as positive identification. DisasterVLG incorporates these diagnostic principles through reasoning-oriented and multiple-choice VQA across open-ended and constrained answer spaces. In the RS domain, models have progressively advanced from basic captioning (RS-GPT~\cite{hu2025rsgpt}) to region-level grounding (GeoChat~\cite{kuckreja2024geochat}, LHRS-Bot~\cite{muhtar2024lhrs}, SkyEyeGPT~\cite{zhan2025skyeyegpt}), and more recently to multi-sensor, multi-temporal reasoning (EarthGPT~\cite{zhang2024earthgpt}, GeoBench-VLM~\cite{danish2025geobench-vlm}, EarthDial~\cite{soni2025earthdial}, EarthMind~\cite{shu2025earthmind}). Yet these remain general-purpose, with disaster coverage limited to isolated tasks like flood detection or damaged building counting. DisasterVLG differentiates itself through comprehensive vertical support for the humanitarian domain, bridging general-purpose RS reasoning and the specialized demands of on-the-ground crisis response. {The region-level annotation complements answer-level evidence by identifying the image regions associated with caption phrases and the entities participating in a stated relationship. These links provide a spatial target for grounding, separate from the quality of the generated answer.}

%% file: sections/dataset.tex
\label{sec:dataset}

Disaster image understanding requires models to recognize hazards, damaged infrastructure, affected environments, response activities, and uncertainty under visually noisy and geographically diverse conditions. DisasterVLG is designed to address these challenges by providing large-scale, evidence-based supervision for disaster-specific vision-language learning, with broad coverage of disaster types, geographic contexts, visual scenes, and humanitarian-relevant information needs. Next, we describe the dataset collection, filtering, and annotation steps.

\subsection{Data Collection and Filtering}
We collected over 7 million images from Twitter/X between 2018 and 2023 using hundreds of disaster-specific keywords, hashtags, event names, and multilingual query terms spanning floods, earthquakes, wildfires, hurricanes, landslides, explosions, and severe storms. After removing approximately 2 million exact and near-duplicate images via hashing and color histogram filtering, the remaining 5 million images were screened with Qwen3-VL-32B~\cite{qwen3vl} for visually observable disaster-related content, yielding the 1.2 million \textit{potentially} relevant images that constitute DisasterVLG.

\begin{figure*}[t]
\centering
\includegraphics[width=0.95\textwidth]{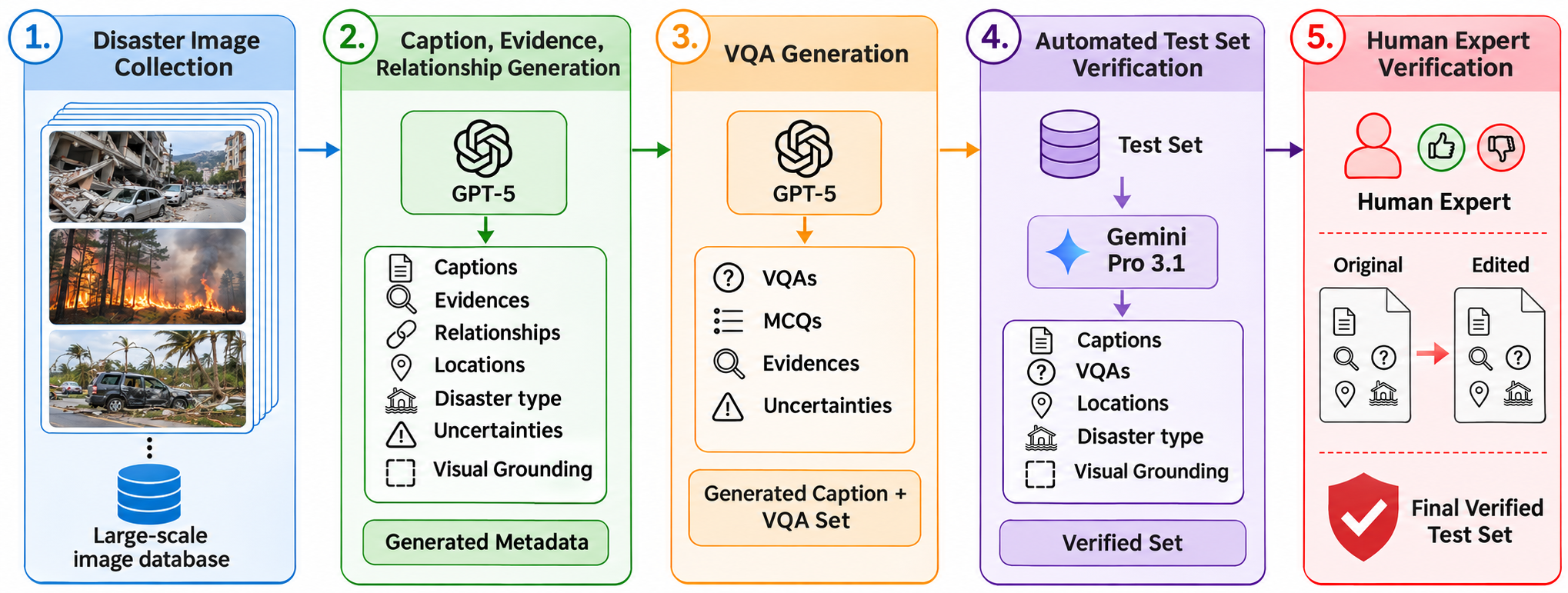}
\caption{DisasterVLG data and annotation pipeline: A curated disaster image dataset is first collected. GPT-5~\cite{gpt-5} then generates image-grounded captions and supporting evidence, followed by tailored VQA annotations conditioned on the caption, incident assessment, supporting evidence, relationships, and uncertainties. Finally, a human and Gemini 3.1 Pro~\cite{gemini31pro} work together to create a verified test set.}
\label{fig:data_model}
\end{figure*}

\subsection{Multi-dimensional Scene Captioning and Evidence Extraction}
As shown in Figure~\ref{fig:data_model}, in the first step, each filtered image was annotated with GPT-5~\cite{gpt-5} to generate: \textit{(i)} a dense factual caption, \textit{(ii)} an incident assessment, \textit{(iii)} visual evidence, \textit{(iv)} a location assessment, \textit{(v)} visible spatial and interaction relationships, and \textit{(vi)} uncertainties. The model was instructed to base all outputs only on visible evidence and to avoid unsupported assumptions about causes, locations, identities, injuries, or intent.

The \textit{caption} describes the main visible content of the image, including damaged infrastructure, environmental conditions, objects, property conditions, human presence, and spatial layout. The \textit{incident assessment} assigns a \textit{likely incident type}, together with a confidence level and provides \textit{visual evidence}. For example, collapsed walls, exposed interiors, rubble, and heavy machinery near a damaged building may support a structural-damage label, while the exact cause may remain unknown. The likely types produced a long-tail distribution of more than 12K incident variants, which we mapped to 29 canonical disaster/hazard types, such as flood, wildfire, road traffic accident.
The \textit{location assessment} records possible geographic cues only when visible evidence is available, such as signs, landmarks, language, or architecture; otherwise, the location is marked as unknown.

The annotation also captures \textit{relationships} among visible entities, such as heavy machinery positioned next to a damaged building, rubble accumulated along a sidewalk, people standing near equipment, or adjacent buildings appearing intact. Readable text is transcribed when present. Finally, \textit{uncertainty} notes identify ambiguous elements that cannot be confirmed visually, such as whether heavy machinery is being used for demolition or debris removal, the extent of unseen damage, or the absence of reliable geographic cues. Prompt templates for this step can be found in Appendix~\ref{app:caption_prompt}.

\subsection{Visual Question-Answer Generation}
We generate \textit{visual question-answer} pairs to help VLMs learn disaster-scene understanding beyond captioning. We aim for diverse VQAs (binary, MCQs, and open-ended) that cover hazard recognition, damage interpretation, spatial reasoning, critical infrastructure, response presence, difficulty level, and uncertainty handling. For each image, GPT-5 is provided with annotations from the last stage: generated caption, incident assessment, supporting evidence, relationships, and uncertainties. The model is then asked to generate 12--15 positive QA pairs and up to five negative yes/no questions.

Positive questions focus on visible, crisis-relevant details, such as hazard type, structural damage, rubble location, proximity to machinery, traffic infrastructure, or possible responder presence. Answers are required to remain image-based, use visible cues, and express uncertainty when needed. Negative questions serve as hallucination checks by asking about absent conditions, such as whether flooding or water damage is visible in a structurally damaged building scene, whether fire or smoke is shown when no burn or smoke cues are present, or whether emergency vehicles such as fire trucks or ambulances appear when none are visible. Prompt templates for VQA and MCQ generation can be found in Appendix~\ref{app:vqa_prompt} and Appendix~\ref{app:mcq_prompt}, respectively.

\subsection{Region-Level Visual Grounding Annotations}
\label{sec:visual_grounding_annotations}
DisasterVLG has a visual grounding layer for the existing captions and relationship statements. Each annotated region consists of a bounding box and a short description of its visible content, such as a tree trunk engulfed in flames. Grounded caption phrases are linked to one or more corresponding regions. These links retain the connection between the original description and its spatial evidence rather than treating the boxes as independent object labels.

Phrases that cannot be assigned a box are retained as unresolved, with a reason indicating an absence statement, non-visual content, a not-visible entity, or an ambiguous reference. For example, ``no people are visible'' is recorded as an absence statement rather than assigned a person box.

\subsection{Test-set Construction}
Next, we design a test set that is reliable, geographically broad, and representative of diverse disaster types. As most images in DisasterVLG were collected during real-world disaster events across various geographies, such as the 2019 Hurricane Dorian and the 2023 Turkey earthquake, the dataset carries natural geographic diversity that we leverage during test set construction. For each canonical disaster category, we set a sampling target of up to 10\% of available instances, capped at 4,000 images per category. To ensure intra-category diversity rather than redundant near-duplicate sampling, we apply k-means clustering within each category prior to sampling, setting the number of clusters equal to the per-category target. One representative image is then drawn from each cluster, with priority given to samples that carry event-based location information, reinforcing geographic breadth in the final selection. As k-means does not always yield the full target number of clusters, the final per-category count may be slightly below the target, resulting in a test set of 38K images. This design enables evaluation of both geographic generalization, such as model performance across regions, and disaster-type generalization, such as robustness across different hazard categories, spanning 75 countries and 28 disaster/hazard types.

 \subsection{Human Verification}
To improve test-set reliability, we audited the test annotations (captions, incident evidence, location assessments, and VQAs) with Gemini 3.1 Pro~\cite{gemini31pro}, followed by human verification. Given each image and its full annotation record, Gemini checked whether every claim was visually supported, internally consistent, and relevant to disaster assessment. It was instructed to keep the original annotation by default and to correct it only when an error was clear (prompt in Appendix~\ref{app:verifier_prompt}). We grouped the proposed edits by magnitude: edits with less than 80\% similarity between the original and revised text were labeled \textit{major}, edits above this threshold \textit{moderate}, and small wording or stylistic suggestions \textit{minor}. Human reviewers inspected 2--5\% of the edits in each group. About 70\% of major edits were valid corrections, compared with fewer than 20\% of moderate edits, and minor edits were generally unnecessary. We therefore accepted only the major edits and kept the original GPT-5 annotations for moderate and minor cases. To estimate the residual error after this audit, we human-validated 1,500 test images: 750 with accepted major edits and 750 drawn from the remaining test images, stratified by canonical type. Two annotators (expert in disaster image analysis) independently judged each canonical label, caption claim, and VQA answer, and a third expert adjudicated disagreements. Inter-annotator agreement was 98.0\% for canonical labels (Cohen's $\kappa=0.98$), 96.0\% for caption claims, and 95.0\% for VQA correctness. After adjudication, 0.9\% of canonical labels (14/1,500), 1.3\% of caption claims (39/3,000), and 2.1\% of answerable VQA answers (65/3,000) in the verification samples needed minor modifications.

\begin{table*}[t]
\centering
\caption{Summary statistics of DisasterVLG for the full dataset, training set, and test set.}
\label{tab:dataset_stats}
\scriptsize
\setlength{\tabcolsep}{2pt}
\resizebox{\textwidth}{!}{%
\begin{tabular}{lccccccc|cccccc}
\toprule
& \multicolumn{7}{c|}{\textbf{Scale}}
& \multicolumn{6}{c}{\textbf{Annotation \& VQA Structure}} \\
\cmidrule(lr){2-8}\cmidrule(lr){9-14}
\textbf{Split}
& \textbf{Images}
& \textbf{CT}
& \textbf{IT}
& \textbf{QA}
& \textbf{Q/I}
& \textbf{MCQ}
& \textbf{VG}
& \textbf{Rel/Unc/Loc}
& \textbf{CapLen}
& \textbf{Pos/Neg}
& \textbf{FF}
& \textbf{E/M/H}
& \textbf{Supp/Abs} \\
\midrule
\textbf{Train}
& \#1,197,351
& 29
& 12K
& 19.9M
& 16.64
& 2.6M
& 7.59M
& 97.5 / 98.0 / 69.04
& 88.78
& 79.8 / 20.2
& 59.6
& 57.6 / 38.4 / 4.0
& 1.56 / 0.34 \\
\textbf{Test}
& 38,254
& 28
& 1,498
& 641K
& 16.76
& 85K
& 350K
& 99.0 / 99.7 / 59.70
& 83.78
& 79.1 / 20.9
& 62.7
& 51.9 / 42.4 / 5.8
& 1.55 / 0.35 \\
\midrule
\textbf{Full}
& 1.24M
& 29
& 12K
& 20.5M
& 16.65
& 2.7M
& 7.89M
& 97.5 / 98.1 / 69.12
& 88.63
& 79.8 / 20.2
& 59.7
& 57.4 / 38.5 / 4.1
& 1.56 / 0.34 \\
\bottomrule
\end{tabular}
}
\par\vspace{2pt}
{\scriptsize \textbf{CT} = canonical type; \textbf{IT} = incident type (i.e., GPT-5 assigned ``likely type"); \textbf{QA} = question-answer pairs; \textbf{Q/I} = questions per image; \textbf{VG} = visual grounding; \\ \textbf{Rel/Unc/Loc} = relationships, uncertainties, and location assessment (\%); \textbf{CapLen} = caption length; \textbf{Pos/Neg} = polarity (\%); \\ \textbf{FF} = free-form (\%); \textbf{E/M/H} = Easy/Medium/Hard (\%); \textbf{Supp/Abs} = support-evidence and absence-check items.}
\end{table*}

\subsection{Dataset Statistics and Coverage}

Table~\ref{tab:dataset_stats} summarizes the scale, coverage, and annotation stats of DisasterVLG. The dataset contains 1,235,605 images across 29 canonical hazard types, with 1.2M detailed evidence-grounded captions, more than 23M question annotations (20.57M QA pairs; 2.71M MCQs) and 7.9M visual grounding bounding boxes. Each image has 16 QA pairs on average. Beyond the canonical taxonomy, incident labels form a much richer long tail of about 12K likely incident types, capturing the fine-grained diversity of real disaster imagery. The test set preserves all 28 canonical types and includes 726,869 question annotations (641,167 QA pairs; 85,702 MCQs) spanning 1,498 likely incident types.

DisasterVLG also provides broad geographic and semantic coverage. Location metadata is available for about 70\% of the full dataset (covering 142 countries) and 60\% of the test set. Captions are dense, averaging 88.63 words, indicating detailed scene descriptions rather than short labels. Most examples include explicit spatial relationships and uncertainty statements (97.5\% and 98.1\% in the full set), supporting grounded reasoning and uncertainty-aware evaluation. The VQA distribution remains consistent across splits, with roughly 80/20 positive-negative polarity, about 60\% free-form answers, and a full-set difficulty mix of 57.4/38.5/4.1\% easy, medium, and hard questions. Additional dataset statistics can be found in Appendix~\ref{app:dataset_stats}.

%% file: sections/tasks_models.tex
\label{sec:tasks_models}

\paragraph{DisasterVLG Tasks.} We define four tasks for both training and evaluation, covering different levels of hazard image understanding: factual scene captioning, visual question answering, incident reasoning, and multiple-choice VQA. Across all tasks, models are expected to ground their outputs in visible image evidence and avoid unsupported inference.

\begin{itemize}[leftmargin=.8em, itemsep=0pt, topsep=0pt, label={}]
    \item \textbf{T1: Disaster scene captioning}: Given an image, the model generates a factual description grounded in visible content. The task uses the caption for training, supported by incident and uncertainty annotations. Captions emphasize disaster-relevant evidence such as infrastructure damage, environmental hazard, and response activity.
    \item \textbf{T2: Grounded VQA reasoning}: Given an image and an open-ended question, the model generates an answer grounded in visible evidence. Each training QA example includes the question, answer, answer type, polarity, and supporting evidence, including checks for absent cues. This task tests whether models can answer disaster-relevant questions, reason from visual cues, and acknowledge when the image does not provide enough evidence.
    \item \textbf{T3: Grounded incident reasoning}: Given an image, the model infers the likely disaster or incident type and justifies it using visual evidence. For training, we use incident and canonical labels, evidence statements, and uncertainty annotations. The goal is not only to classify the incident, but to explain why the label is supported and when the image does not justify a stronger claim.
    \item \textbf{T4: Multiple-choice VQA}: Given an image, a question, and candidate answers, the model selects the correct option. The task is derived from grounded QA and incident annotations, using plausible distractors to test whether models can distinguish visually supported answers from similar but unsupported alternatives. It provides a controlled evaluation of disaster-specific visual reasoning.
    \item {\textbf{T5: Region-level visual grounding}: Given an image, the model localizes disaster-relevant evidence with bounding boxes on a 0--1000 normalized grid, in two settings. In \textit{detection}, it receives only the image and returns a labeled box for every disaster-relevant object and region; the targets are the annotated regions. In \textit{grounded captioning}, it writes a detailed description and grounds its key phrases with boxes in the same output; the targets are the regions linked to phrases of the reference caption. Detection is scored solely on box placement; grounded captioning additionally requires each box to contain the matching phrase (Sec.~\ref{sec:visual_grounding_evaluation}).}
\end{itemize}

\paragraph{DisasterVLG Models.} We develop a family of disaster-specialized vision-language models, DisasterVLMs, built on the Qwen3.5~\cite{qwen35blog} model family. Qwen3.5 serves as a strong foundation due to its competitive multimodal reasoning capabilities, open availability, and support for high-resolution visual inputs --- properties particularly relevant for the fine-grained scene understanding required in crisis imagery. We train 2B, 4B, and 9B variants; the main results (Table~\ref{tab:merged_results}) report the 4B and 9B models, and the 2B model is included in the scaling analysis (Appendix~\ref{app:model_scale_data_size}).

Rather than full fine-tuning, we adopt parameter-efficient adaptation via LoRA~\cite{hu2022lora}, applying low-rank updates to both the language model and the multimodal projector while keeping the vision backbone frozen. This concentrates trainable capacity on cross-modal alignment and task-specific reasoning, mitigates the risk of catastrophic forgetting of pretrained visual representations, and substantially reduces memory requirements. We use LoRA rank 32, alpha 64, and dropout 0.05, with 16-bit mixed-precision training.

All models are trained jointly across the five DisasterVLG tasks using instruction-formatted inputs, with task identity encoded via prompt templates. Training is conducted on 4 H200 GPUs for 2 epochs using a learning rate of $3\times10^{-5}$, 100 warmup steps, and a maximum sequence length of 1024, with an effective batch size of 640 via gradient accumulation. Gradient checkpointing and FlashAttention-2 are enabled throughout for memory-efficient, stable training. At inference time, models are evaluated in a zero-shot prompted setting using the same instruction templates as training, without any task-specific decoding adjustments.

%% file: sections/experiments.tex
\label{sec:experiment}

\subsection{Baselines and Metrics}
\label{sec:baselines_metrics}

We compare DisasterVLMs with ten zero-shot baselines from six model families: Molmo2~\cite{molmo2}, InternVL3.5-8B~\cite{internvl35}, Qwen3-VL (4B, 8B, 32B)~\cite{qwen3vl}, Gemma4-4B~\cite{gemma4}, the Qwen3.5 base models (2B, 4B, 9B)~\cite{qwen35blog} from which DisasterVLMs are adapted, and the proprietary GPT-5~\cite{gpt-5} as a strong closed-model reference. All baselines use the same inference prompts as DisasterVLM, without task-specific fine-tuning. Table~\ref{tab:merged_results} reports the 4B and 9B DisasterVLM variants.

For the generative tasks, we report ROUGE-L~\cite{Rouge}, which measures lexical overlap with the reference, and Sentence-T5 Sharpened Cosine Similarity (ST5-SCS)~\cite{rscc}, which measures semantic similarity in embedding space and therefore credits valid answers that differ lexically from the reference. We also report GES, described below. Visual grounding (T5) is scored with box-level F1 (Sec.~\ref{sec:visual_grounding_evaluation}). For incident reasoning, we additionally report incident-type accuracy, and for multiple-choice VQA, exact answer accuracy.

{\paragraph{Reference-grounded evaluation.}
For grounded evaluation, we additionally use the Grounded Evidence Score ($\mathrm{GES}$), which measures factual agreement with the reference caption and answers. An LLM-based aligner labels each extracted response claim as matched, contradicted, or not in the reference. Reference precision is the fraction of claims matched to reference facts; recall is the fraction of distinct reference facts covered by matched claims. Their harmonic mean gives $\mathrm{GES}$ similar to F1 score. The strict score penalizes all unmatched claims, including visually correct details but not relevant to disaster and not in the reference, and therefore measures reference alignment rather than visual correctness alone. Appendix~\ref{app:ges_ref} gives the full definition, scoring variants, and aggregation protocol.
}

\subsection{Experimental Results}

Table~\ref{tab:merged_results} summarizes results on all five tasks (T1--T5). \texttt{DisasterVLM-9B} obtains the best score in every column except MCQ accuracy, where \texttt{DisasterVLM-4B} is best and the 9B model is on par with GPT-5. The margin over the strongest baseline depends on the metric: it is large for ROUGE-L on all open-ended tasks and for ST5-SCS on VQA and incident reasoning, and much smaller for GES and the grounding F1 scores. The 4B variant does not consistently outperform GPT-5: it trails on GES for captioning and VQA, on both grounding metrics, and on incident accuracy.

\paragraph{Disaster Scene Captioning.}
\texttt{DisasterVLM-9B} achieves the best captioning scores on all three metrics (40.51 ROUGE-L, 89.25 ST5-SCS, 69.4 GES), compared with 31.34, 87.67, and 64.8 for GPT-5, the strongest baseline on each. \texttt{DisasterVLM-4B} also exceeds every baseline on ROUGE-L (37.78) and ST5-SCS (88.53), but trails GPT-5 on GES (62.1 vs.\ 64.8), so its lexical advantage does not translate into higher factual agreement with the reference. More generally, the ROUGE-L gain of the 9B model (+9.17) is about twice its GES gain (+4.6), which suggests that part of the improvement reflects alignment with the reference caption style.

\paragraph{Grounded VQA-Reasoning.}
This task shows the largest gains on the reference-overlap metrics. \texttt{DisasterVLM-9B} reaches 44.18 ROUGE-L and 93.16 ST5-SCS, against 26.45 (GPT-5) and 81.20 (Qwen3-VL-32B) for the strongest baselines, and \texttt{DisasterVLM-4B} is close behind (42.72 / 92.07). On GES the gap narrows: the 9B model exceeds GPT-5 by 2.3 points (65.7 vs.\ 63.4), and the 4B model is below it (59.6). The baselines' answers therefore differ from the references much more in wording than in GES, which scores claim support and answer correctness rather than surface overlap.

\begin{table*}[t]
\centering
\caption{Results on captioning, visual grounding, VQA reasoning, incident reasoning, and multiple-choice QA (MCQ). Column leaders are in \textbf{bold}. Dashes indicate missing scores.}
\label{tab:merged_results}
\small
\setlength{\tabcolsep}{3.5pt}
\begin{tabular}{l rrr|rr|rrr|rrrr|r}
\toprule
& \multicolumn{3}{c|}{\textbf{Captioning}}
& \multicolumn{2}{c|}{\textbf{Visual Grounding}}
& \multicolumn{3}{c|}{\textbf{VQA Reasoning}}
& \multicolumn{4}{c|}{\textbf{Incident Reasoning}}
& \multicolumn{1}{c}{\textbf{MCQ}} \\
\cmidrule(lr){2-4}\cmidrule(lr){5-6}\cmidrule(lr){7-9}\cmidrule(lr){10-13}\cmidrule(lr){14-14}
\textbf{Model}
& \multicolumn{1}{c}{\textbf{R-L}}
& \multicolumn{1}{c}{\textbf{ST5}}
& \multicolumn{1}{c|}{\textbf{GES}}
& \multicolumn{1}{c}{\textbf{\begin{tabular}[b]{@{}c@{}}D-F1\end{tabular}}}
& \multicolumn{1}{c|}{\textbf{\begin{tabular}[b]{@{}c@{}}GC-F1\end{tabular}}}
& \multicolumn{1}{c}{\textbf{R-L}}
& \multicolumn{1}{c}{\textbf{ST5}}
& \multicolumn{1}{c|}{\textbf{GES}}
& \multicolumn{1}{c}{\textbf{R-L}}
& \multicolumn{1}{c}{\textbf{ST5}}
& \multicolumn{1}{c}{\textbf{GES}}
& \multicolumn{1}{c|}{\textbf{Acc}}
& \multicolumn{1}{c}{\textbf{Acc}} \\
\midrule
Molmo2~\cite{molmo2} & \grad{8}{23.89} & \grad{11}{81.90} & \grad{15}{52.4} & \grad{15}{46.4} & \grad{12}{39.6} & \grad{9}{17.96} & \grad{8}{75.24} & \grad{16}{53.9} & \grad{13}{21.25} & \grad{11}{73.64} & \grad{12}{55.2} & \grad{16}{70.84} & \grad{14}{63.57} \\
InternVL3.5-8B~\cite{internvl35} & \grad{12}{25.12} & \grad{22}{83.45} & \grad{26}{56.7} & \grad{33}{53.6} & \grad{31}{47.5} & \grad{11}{19.09} & \grad{16}{77.89} & \grad{27}{56.8} & \grad{15}{22.87} & \grad{19}{76.59} & \grad{17}{56.9} & \grad{23}{72.60} & \grad{21}{65.83} \\
Qwen3-VL-4B~\cite{qwen3vl} & \grad{8}{23.98} & \grad{20}{83.20} & \grad{20}{54.5} & \grad{17}{47.2} & \grad{19}{42.5} & \grad{8}{17.80} & \grad{10}{75.83} & \grad{17}{54.1} & \grad{11}{19.78} & \grad{16}{75.59} & \grad{13}{55.6} & \grad{12}{69.92} & \grad{10}{62.44} \\
Qwen3-VL-8B~\cite{qwen3vl} & \grad{9}{24.29} & \grad{20}{83.15} & \grad{25}{56.2} & \grad{31}{52.5} & \grad{26}{45.1} & \grad{10}{18.73} & \grad{14}{77.25} & \grad{15}{53.6} & \grad{10}{19.16} & \grad{12}{73.78} & \grad{14}{55.7} & \grad{15}{70.61} & \grad{12}{63.14} \\
Qwen3.5-2B~\cite{qwen35blog} & \grad{9}{24.26} & \grad{8}{81.40} & \grad{8}{49.8} & \grad{8}{43.5} & \grad{8}{37.9} & \grad{8}{17.61} & \grad{10}{75.78} & \grad{8}{51.7} & \grad{8}{18.05} & \grad{8}{72.22} & \grad{8}{53.5} & \grad{8}{68.85} & \grad{8}{61.73} \\
Qwen3.5-4B~\cite{qwen35blog} & \grad{9}{24.30} & \grad{17}{82.71} & \grad{24}{55.7} & \grad{23}{49.6} & \grad{24}{44.4} & \grad{9}{18.10} & \grad{12}{76.71} & \grad{22}{55.6} & \grad{16}{23.03} & \grad{15}{75.24} & \grad{13}{55.5} & \grad{18}{71.40} & \grad{16}{64.32} \\
Qwen3.5-9B~\cite{qwen35blog} & \grad{12}{25.25} & \grad{16}{82.56} & \grad{29}{57.8} & \grad{38}{55.3} & \grad{35}{48.8} & \grad{11}{19.17} & \grad{16}{78.08} & \grad{31}{57.8} & \grad{16}{23.13} & \grad{21}{77.44} & \grad{18}{57.3} & \grad{30}{74.20} & \grad{26}{67.37} \\
Gemma4-4B~\cite{gemma4} & \grad{10}{24.65} & \grad{18}{82.90} & \grad{29}{57.6} & -- & -- & \grad{11}{19.05} & \grad{14}{77.24} & \grad{21}{55.3} & \grad{16}{23.10} & \grad{18}{76.45} & \grad{10}{54.2} & \grad{26}{73.28} & \grad{52}{75.79} \\
Qwen3-VL-32B~\cite{qwen3vl} & \grad{14}{25.82} & \grad{45}{87.04} & \grad{39}{61.5} & \grad{46}{58.5} & \grad{44}{52.7} & \grad{23}{25.45} & \grad{25}{81.20} & \grad{45}{61.7} & \grad{22}{27.21} & \grad{27}{80.10} & \grad{31}{62.4} & \grad{34}{75.24} & \grad{53}{76.12} \\
GPT-5~\cite{gpt-5} & \grad{31}{31.34} & \grad{50}{87.67} & \grad{48}{64.8} & \grad{57}{62.8} & \grad{52}{55.9} & \grad{25}{26.45} & \grad{24}{80.76} & \grad{51}{63.4} & \grad{26}{29.65} & \grad{33}{82.43} & \grad{49}{69.2} & \grad{49}{78.85} & \grad{58}{77.73} \\
\midrule
DisasterVLM-4B & \grad{51}{37.78} & \grad{55}{88.53} & \grad{41}{62.1} & \grad{54}{61.6} & \grad{49}{54.7} & \grad{57}{42.72} & \grad{57}{92.07} & \grad{37}{59.6} & \grad{50}{45.23} & \grad{51}{89.84} & \grad{53}{70.9} & \grad{45}{77.93} & \grad{60}{\textbf{78.22}} \\
\textbf{DisasterVLM-9B} & \grad{60}{\textbf{40.51}} & \grad{60}{\textbf{89.25}} & \grad{60}{\textbf{69.4}} & \grad{60}{\textbf{64.1}} & \grad{60}{\textbf{59.2}} & \grad{60}{\textbf{44.18}} & \grad{60}{\textbf{93.16}} & \grad{60}{\textbf{65.7}} & \grad{60}{\textbf{51.50}} & \grad{60}{\textbf{93.70}} & \grad{60}{\textbf{73.6}} & \grad{60}{\textbf{81.52}} & \grad{58}{77.70} \\
\midrule
\textbf{$\Delta$ vs.\ best baseline} & \gain{+9.17} & \gain{+1.58} & \gain{+4.6} & \gain{+1.3} & \gain{+3.3} & \gain{+17.73} & \gain{+11.96} & \gain{+2.3} & \gain{+21.85} & \gain{+11.27} & \gain{+4.4} & \gain{+2.67} & \gain{+0.49} \\
\bottomrule
\end{tabular}
\par\vspace{2pt}
{\scriptsize \textbf{R-L}: ROUGE-L; \textbf{ST5}: ST5-SCS; \textbf{GES}: Grounded Evidence Score (Sec.~\ref{sec:baselines_metrics}); \textbf{Acc}: accuracy; {\textbf{D-F1}: detection box F1 at IoU\,$\geq$\,0.5; \textbf{GC-F1}: grounded-captioning F1, requiring IoU\,$\geq$\,0.5 and phrase agreement (Sec.~\ref{sec:visual_grounding_evaluation}).} Where shown, cell shading is normalized per column; darker blue indicates higher performance. $\Delta$: per-column difference between the best DisasterVLM variant and the best baseline.}
\end{table*}

\paragraph{Grounded Incident Reasoning.}
This task requires both correct incident recognition and a grounded visual justification. \texttt{DisasterVLM-9B} is best on all four metrics (51.50 ROUGE-L, 93.70 ST5-SCS, 73.6 GES, 81.52 accuracy), against 29.65, 82.43, 69.2, and 78.85 for GPT-5. \texttt{DisasterVLM-4B} exceeds GPT-5 on the three generative metrics, including GES (70.9 vs.\ 69.2), but not on incident accuracy (77.93 vs.\ 78.85). The accuracy gain of the 9B model (+2.67) is much smaller than its gains in explanation overlap, which suggests that adaptation mainly changes how incidents are justified rather than whether they are identified.

\paragraph{Multiple-Choice VQA.}
MCQ shows the smallest differences between adapted and general-purpose models. \texttt{DisasterVLM-4B} is best (78.22), 0.49 points above GPT-5 (77.73), and \texttt{DisasterVLM-9B} (77.70) is on par with GPT-5. Both adapted models improve substantially over their Qwen3.5 base models (64.32 and 67.37, gains of 13.90 and 10.33 points), but Qwen3-VL-32B (76.12) and Gemma4-4B (75.79) are also within 2.5 points of the best result. The constrained answer format therefore appears to reduce the advantage of domain-specific supervision relative to open-ended generation. The 0.52-point difference between the 4B and 9B variants is too small to rank them without confidence intervals.
\paragraph{Visual Grounding.}
\label{sec:visual_grounding_evaluation}
Both T5 evaluations are scored with box-level F1 under greedy one-to-one matching at a minimum IoU of 0.5. True positives, predicted boxes and reference boxes are summed over the test set before precision, recall and F1 are computed, so an image without predictions contributes all of its references as misses. \textit{Detection F1} (D-F1) matches predictions to the annotated regions in order of descending IoU. Box labels are not used, so D-F1 measures where the model places boxes, not what it calls them. \textit{Grounded-captioning F1} (GC-F1) uses one reference entry per phrase--region link of the reference caption. A prediction can match an entry only if its phrase also has a cosine similarity of at least 0.5 to the entry's phrase, using Qwen3-Embedding-0.6B; candidates are accepted in order of phrase similarity, with IoU breaking ties. A correct box under the wrong phrase is therefore not credited, and a region described by several phrases must be grounded once under each of them. GC-F1 thus measures phrase grounding rather than localization alone.
\texttt{DisasterVLM-9B} achieves the best grounding scores (64.1 D-F1, 59.2 GC-F1), ahead of GPT-5 (62.8 / 55.9), while \texttt{DisasterVLM-4B} is slightly below GPT-5 on both (61.6 / 54.7).

\subsection{Additional Analyses}
Beyond aggregate task performance, we conduct a series of targeted analyses to better understand model behavior along several dimensions. {In the main paper, we examine structured visual grounding through object- and scene-based term recognition, fine-grained classification behavior across incident types, and geographic generalization across regions.}{In the main paper, we examine object- and scene-based term recognition; incident-level classification behavior and geographic generalization across regions are analyzed in Appendix~\ref{app:grounding_analysis}. Term recognition measures agreement with annotated vocabulary; it does not measure the region localization introduced in T5.} We additionally provide scaling and data efficiency experiments as well as a deeper grounding analysis of object- and scene-level evidence across disaster types in Appendix~\ref{app:model_scale_data_size} and Appendix~\ref{app:grounding_analysis}, respectively.

\begin{table}[t]
\centering
\scriptsize
\caption{Results on object and scene classification}
\label{tab:mcq_results}
\begin{tabular}{@{}lcc@{}}
\toprule
\textbf{Model} & \makecell{\textbf{Object}\\\textbf{Macro-F1}} & \makecell{\textbf{Scene}\\\textbf{Macro-F1}} \\
\midrule
Qwen3.5-9B              & 73.6 & 76.1 \\
DisasterVLM-4B          & 77.9 & 78.2 \\
\textbf{DisasterVLM-9B} & \textbf{81.5} & \textbf{82.85} \\
\bottomrule
\end{tabular}
\end{table}

\paragraph{Object- and Scene-Based Analysis.}

{We evaluate structured visual grounding by extracting disaster-relevant objects and scene terms from reference captions and model predictions, then comparing predicted terms against annotated references.}{We evaluate object and scene term recognition by extracting disaster-relevant terms from reference captions and model predictions, then comparing the predicted terms against annotated references.} This yields separate object-level and scene-level macro-F1 scores measuring the model's ability to recognize relevant object and scene terms. As shown in Table~\ref{tab:mcq_results}, \texttt{DisasterVLM-9B} achieves the strongest performance on both metrics, outperforming both \texttt{DisasterVLM-4B} and the \texttt{Qwen3.5-9B} baseline. {These results indicate that domain adaptation improves not only generative reasoning but also fine-grained grounding of visual entities and scene context under discriminative evaluation.}{These results indicate improved recognition of disaster-relevant object and scene terms.}

\subsection{Discussion}

\texttt{DisasterVLM-9B} outperforms all general-purpose baselines, including GPT-5, on every metric in T1--T5 except MCQ accuracy, where it is on par with GPT-5 (77.70 vs.\ 77.73) and \texttt{DisasterVLM-4B} is best (78.22). Margins are largest on ROUGE-L and ST5-SCS for open-ended and reasoning tasks (up to +21.85 ROUGE-L) and much smaller on GES (+2.3 to +4.6), so part of the gain likely reflects alignment with the reference style rather than additional visual evidence. The 4B variant trails GPT-5 on several reference-grounded and grounding metrics, so the advantage over strong proprietary models is not uniform across scales. Incident-level and cross-continent analyses (Appendix~\ref{app:grounding_analysis}) show errors concentrated among visually similar damage types (\textit{earthquake}, \textit{urban structural damage}, \textit{road traffic accident}) and uneven transfer across regions, pointing to where additional data collection is most needed.

%% file: sections/conclusion.tex
\label{sec:conclusion}
We introduced \textbf{DisasterVLG}, a dataset and benchmark for disaster scene understanding comprising more than 1.24M images from 142 countries across 29 hazard types, with evidence-grounded captions, more than 23M question annotations, and a visual grounding layer that links caption phrases and relationship endpoints to image regions. Beyond identifying the visible hazard, the annotations record the supporting evidence, what remains uncertain, and what cannot be inferred from the image alone. \texttt{DisasterVLM-9B} outperforms all general-purpose baselines, including GPT-5, on captioning, visual grounding, grounded VQA, and incident reasoning, with the largest margins on reference-overlap metrics for reasoning tasks; on multiple-choice QA, adapted and strong general-purpose models perform similarly. Visually overlapping incident types, regional variation, and uneven cross-continent generalization remain open challenges for deployment. Future work will broaden geographic and rare-hazard coverage, strengthen human validation, and extend DisasterVLG toward geospatial grounding and event-level reasoning.

\section*{LLM Usage Statement}
Large language models and vision--language models are used in this work both as
components of the dataset creation and evaluation pipeline described in the paper and for text editing purposes:
\begin{itemize}\itemsep0pt
  \item \textbf{GPT-5} generated the annotation layer of the dataset: dense factual
        captions, incident type and supporting evidence, spatial relationship
        statements, uncertainty statements, open-ended and binary question--answer
        pairs, and multiple-choice questions (Sec.~\ref{sec:dataset}). All released
        annotations derive from this process; the images themselves are collected,
        not generated.
  \item \textbf{Gemini 3.1 Pro} audited the test-split annotations for visual support,
        internal consistency and relevance. Its proposed edits were filtered by edit
        magnitude and verified by human annotators; only major edits were accepted
        (Sec.~\ref{sec:dataset}, Appendix~\ref{app:verifier_prompt}).
  \item \textbf{Qwen3.5 (2B, 4B, 9B)} are the open-weight base models fine-tuned to
        produce the DisasterVLM family (Sec.~\ref{sec:tasks_models}).
  \item Molmo2, InternVL3.5, Qwen3-VL, Gemma4 and GPT-5 are evaluated as zero-shot
        baselines (Sec.~\ref{sec:experiment}). Sentence-T5 and Qwen3-Embedding-0.6B
        embeddings are used in evaluation scoring.
\end{itemize}
\paragraph{Writing.}
The authors take full responsibility for all content of this paper and for the
released dataset, including the correctness of retained model-generated annotations.

%% file: sections/appendices.tex
\section{Additional Dataset Statistics}
\label{app:dataset_stats}

\begin{figure*}[t]
    \centering
    \includegraphics[width=\textwidth]{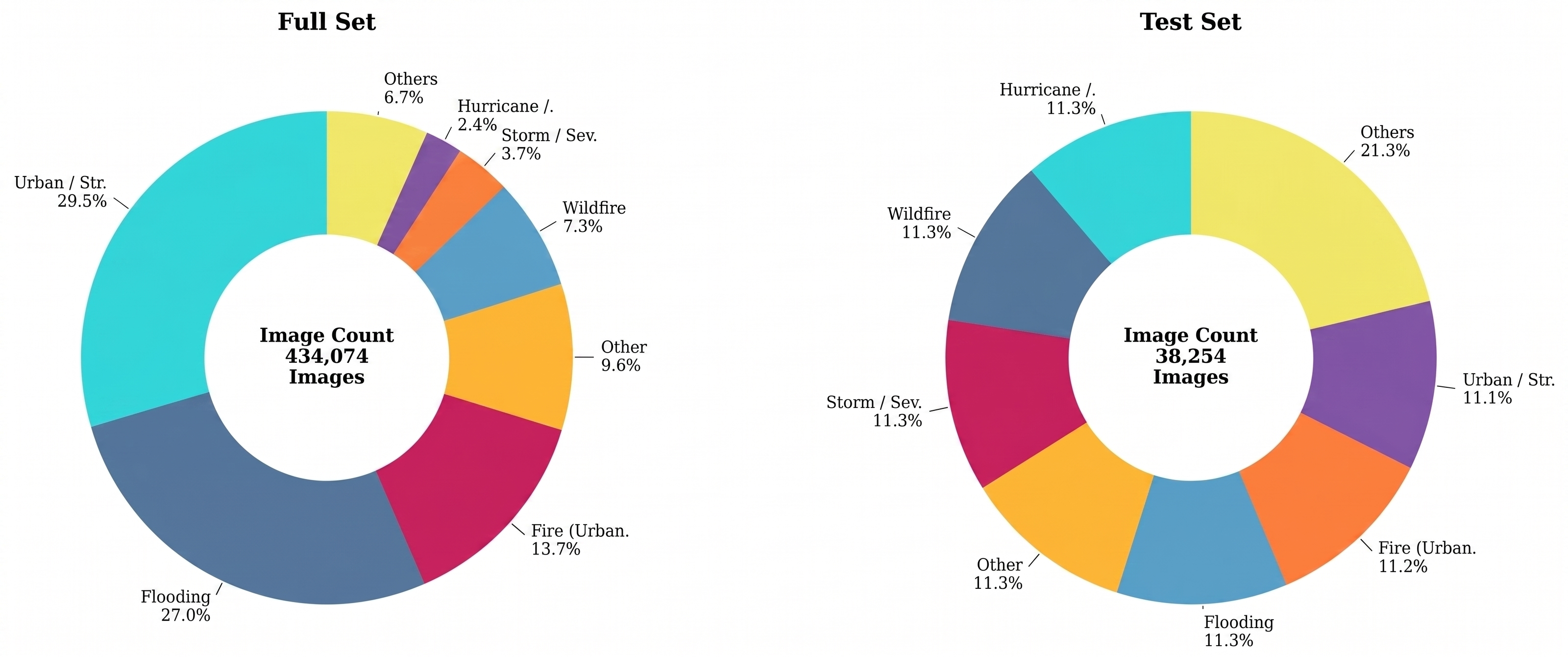}
    \caption{Category distribution of DisasterVLG across the full set and test set. The full set follows the naturally imbalanced distribution of collected crisis imagery, with categories such as \textit{Urban / Structural Damage} and \textit{Flooding} appearing most frequently. In contrast, the test split is constructed to be more balanced across major disaster types, improving the reliability of evaluation across categories. For readability, low-frequency categories are grouped under \textit{Others}. For better insight and comparison, the \textit{Unknown} and \textit{Not-Crisis} categories are removed from the full set distribution.}
    \label{fig:dataset_distribution}
\end{figure*}

Table~\ref{tab:dataset_stats_additional} and Fig~\ref{fig:dataset_distribution} provide a more detailed statistical summary of DisasterVLG for the full dataset, train set, and the final diversity-aware test set.
Beyond the dataset's overall scale, the table highlights the richness of the annotation structure, including relationships, uncertainty fields, visible text, location assessment, and evidence-grounded VQA support.
It also summarizes the distribution of QA types and difficulty levels, showing that the test set remains broadly aligned with the full dataset while being slightly more challenging overall.
These statistics complement the main paper summary by documenting the scope, density, and structural properties of the released benchmark.

\begin{table*}[t]
\centering
\caption{Detailed statistics of DisasterVLG for the full dataset, training set, and test set.}
\label{tab:dataset_stats_additional}
\small
\setlength{\tabcolsep}{5pt}
\begin{tabular}{lccc}
\toprule
Statistic & Full dataset & Train set & Test set \\
\midrule
Total images & 1,235,605 & 1,197,351 & 38,254 \\
Canonical categories & 29 & 29 & 28 \\
Total QA pairs & 20,568,496 & 19,927,329 & 641,167 \\
Average QA pairs / image & 16.65 & 16.64 & 16.76 \\
Median QA pairs / image & 17 & 17 & 17 \\
Total Visual Grounding Boxes & 7,948,308 & 7,597,459 & 350,849 \\
Total MCQs & 2,712,660 & 2,626,958 & 85,702 \\
Average MCQs / image & 2.20 & 2.19 & 2.24 \\
Images with relationships (\%) & 97.5 & 97.5 & 99.0 \\
Images with uncertainties (\%) & 98.1 & 98.0 & 99.7 \\
Images with visible text (\%) & 51.6 & 51.8 & 45.4 \\
Images with known location assessment (\%) & 69.12 & 69.42 & 59.7 \\
Images with collection metadata (\%) & 99.7 & 99.7 & 100.0 \\
Average caption length (words) & 88.63 & 88.78 & 83.78 \\
Positive QA pairs (\%) & 79.8 & 79.8 & 79.1 \\
Negative QA pairs (\%) & 20.2 & 20.2 & 20.9 \\
Free-form QA pairs (\%) & 59.7 & 59.6 & 62.7 \\
Yes/No QA pairs (\%) & 38.6 & 38.7 & 35.6 \\
Easy QA pairs (\%) & 57.4 & 57.6 & 51.9 \\
Medium QA pairs (\%) & 38.5 & 38.4 & 42.4 \\
Hard QA pairs (\%) & 4.1 & 4.0 & 5.8 \\
Avg.\ support evidence items / QA & 1.56 & 1.56 & 1.55 \\
Avg.\ absence-check items / QA & 0.34 & 0.34 & 0.35 \\
\bottomrule
\end{tabular}
\end{table*}

\begin{figure*}[t]
    \centering
    \includegraphics[width=\textwidth]{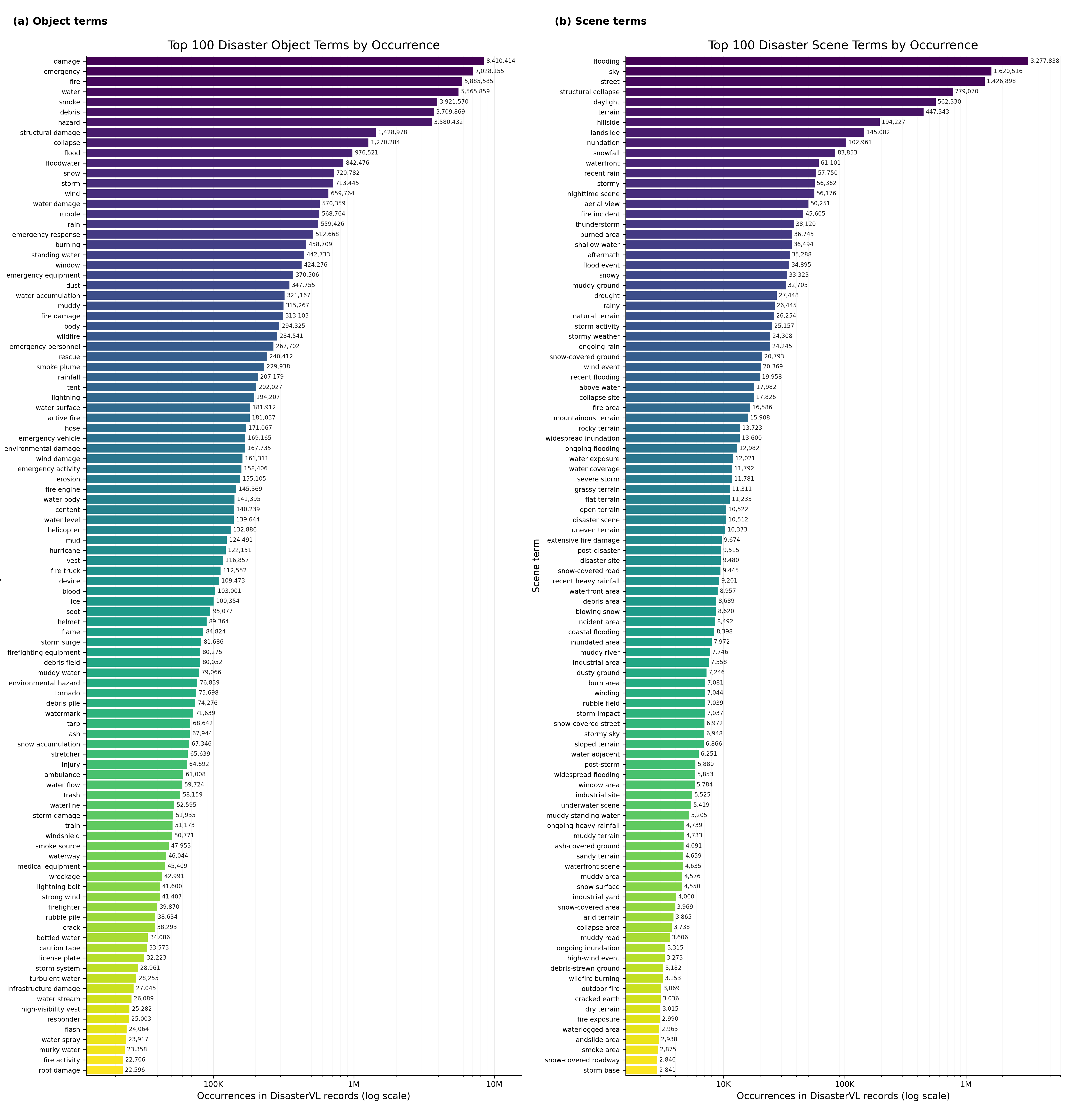}
    \caption{\textbf{Left:} Top 100 disaster-related object terms by occurrence in DisasterVLG records. Frequent terms such as \textit{damage}, \textit{emergency}, \textit{fire}, \textit{water}, \textit{smoke}, and \textit{debris} reflect the strong presence of hazard cues, impacted infrastructure, and response-related entities in the annotations. \textbf{Right:} Top 100 disaster scene terms by occurrence. Common terms such as \textit{flooding}, \textit{street}, \textit{structural collapse}, \textit{daylight}, and \textit{terrain} highlight the dataset's emphasis on broader environmental and situational context. Both panels use a logarithmic x-axis to illustrate the naturally long-tailed distribution of the annotation vocabulary in DisasterVLG.}
    \label{fig:top100_object_scene_terms}
\end{figure*}

\begin{figure*}[t]
    \centering
    \includegraphics[width=0.9\textwidth]{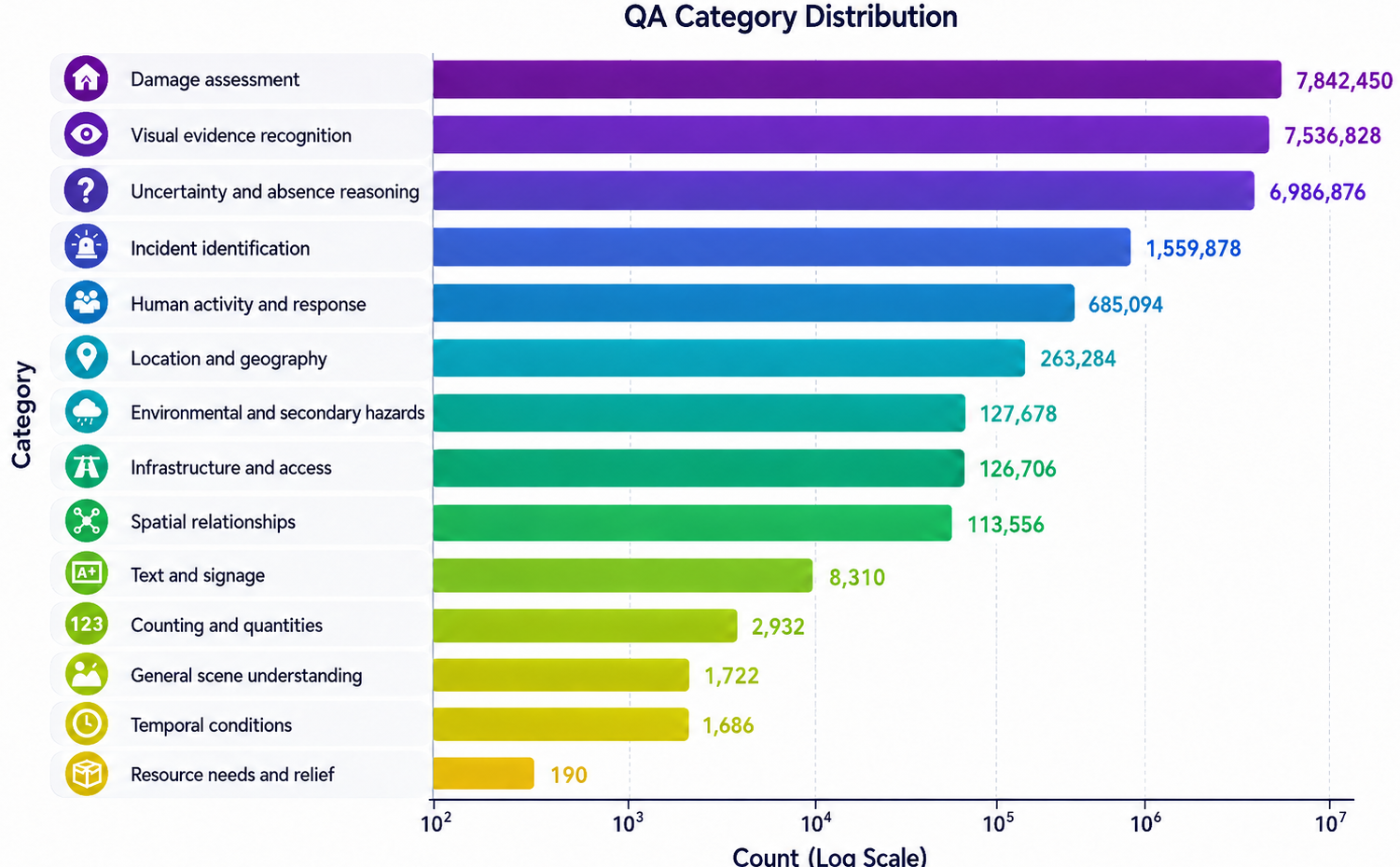}
    \caption{Distribution of VQA categories in DisasterVLG. Counts are shown on a logarithmic scale. The most frequent categories focus on damage assessment, visual evidence recognition, and reasoning about uncertainty or absence, while lower-frequency categories cover more specialized reasoning dimensions such as text recognition, counting, temporal conditions, and resource needs.}
    \label{fig:qa_category_distribution}
\end{figure*}

\subsection{Object and Scene Term Distributions}
\label{app:object_scene_terms}

Figure~\ref{fig:top100_object_scene_terms} provides a lexical summary of the most frequent object and scene terms appearing in DisasterVLG annotations.
The left panel shows the top 100 disaster-related object terms, while the right panel shows the top 100 scene terms, both ranked by occurrence on a logarithmic scale.
These distributions reflect the dense and operationally oriented nature of the dataset: common object/attribute terms such as \textit{damage}, \textit{emergency}, \textit{fire}, \textit{water}, \textit{smoke}, and \textit{debris} indicate that annotations consistently capture visible hazards, affected infrastructure, response-related entities, and physical evidence of disaster impact.
Likewise, frequent scene terms such as \textit{flooding}, \textit{street}, \textit{structural collapse}, \textit{daylight}, \textit{terrain}, and \textit{landslide} show that the dataset also encodes broader environmental and situational context rather than isolated object mentions alone.

The long-tailed shape of both distributions is also informative.
A relatively small number of highly recurrent terms accounts for a large portion of the corpus, while many lower-frequency terms capture more specific disaster cues, terrain conditions, and operational contexts.
This pattern is expected in a large-scale real-world disaster dataset and supports the claim that DisasterVLG contains both broad coverage of common hazards and finer-grained vocabulary for more specialized scenarios.
Together, these statistics provide an additional view of annotation richness beyond image counts and QA totals, showing that the dataset spans a wide spectrum of disaster objects, environments, and scene-level descriptors.

\subsection{Question Category Distribution}
\label{app:qa_category_distribution}

Figure~\ref{fig:qa_category_distribution} summarizes the distribution of VQA categories in DisasterVLG on a logarithmic scale.
The most frequent categories are \textit{damage assessment}, \textit{visual evidence recognition}, and \textit{uncertainty and absence reasoning}, reflecting the dataset's emphasis on operational disaster understanding rather than generic image QA.
These categories directly support tasks that require models to identify visible damage, ground answers in scene evidence, and distinguish observable facts from unsupported inference.

The distribution also shows broad coverage across incident identification, human activity and response, location and geography, environmental hazards, infrastructure access, and spatial relationships.
Lower-frequency categories such as \textit{text and signage}, \textit{counting and quantities}, \textit{temporal conditions}, and \textit{resource needs and relief} capture more specialized dimensions of reasoning.

\section{Scaling Behavior and Data Efficiency}
\label{app:model_scale_data_size}

\begin{figure*}[t]
    \centering
    \begin{subfigure}[t]{0.49\textwidth}
        \centering
        \includegraphics[width=\linewidth]{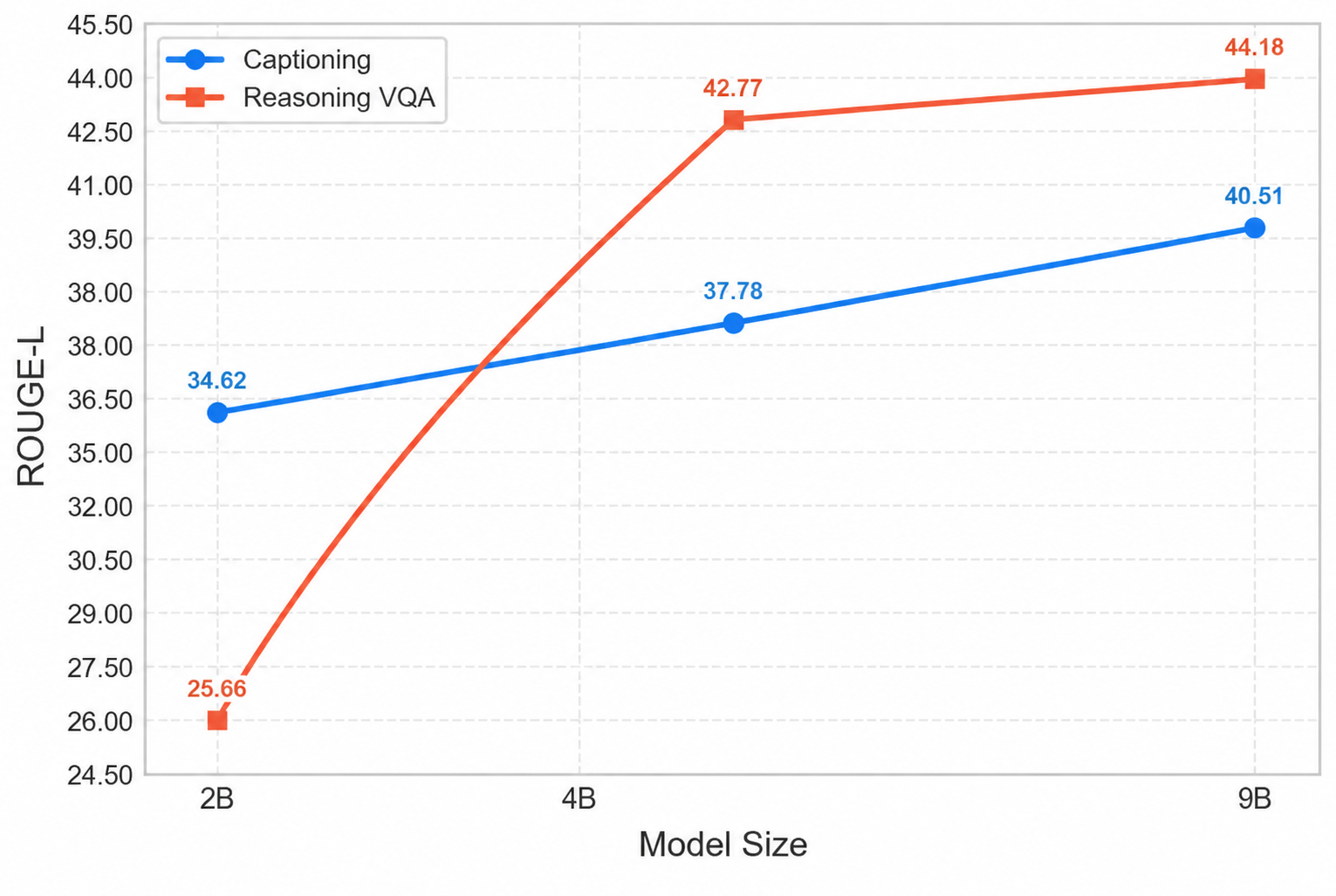}
        \caption{}
        \label{fig:rougel_model_comparison}
    \end{subfigure}
    \hfill
    \begin{subfigure}[t]{0.49\textwidth}
        \centering
        \includegraphics[width=\linewidth]{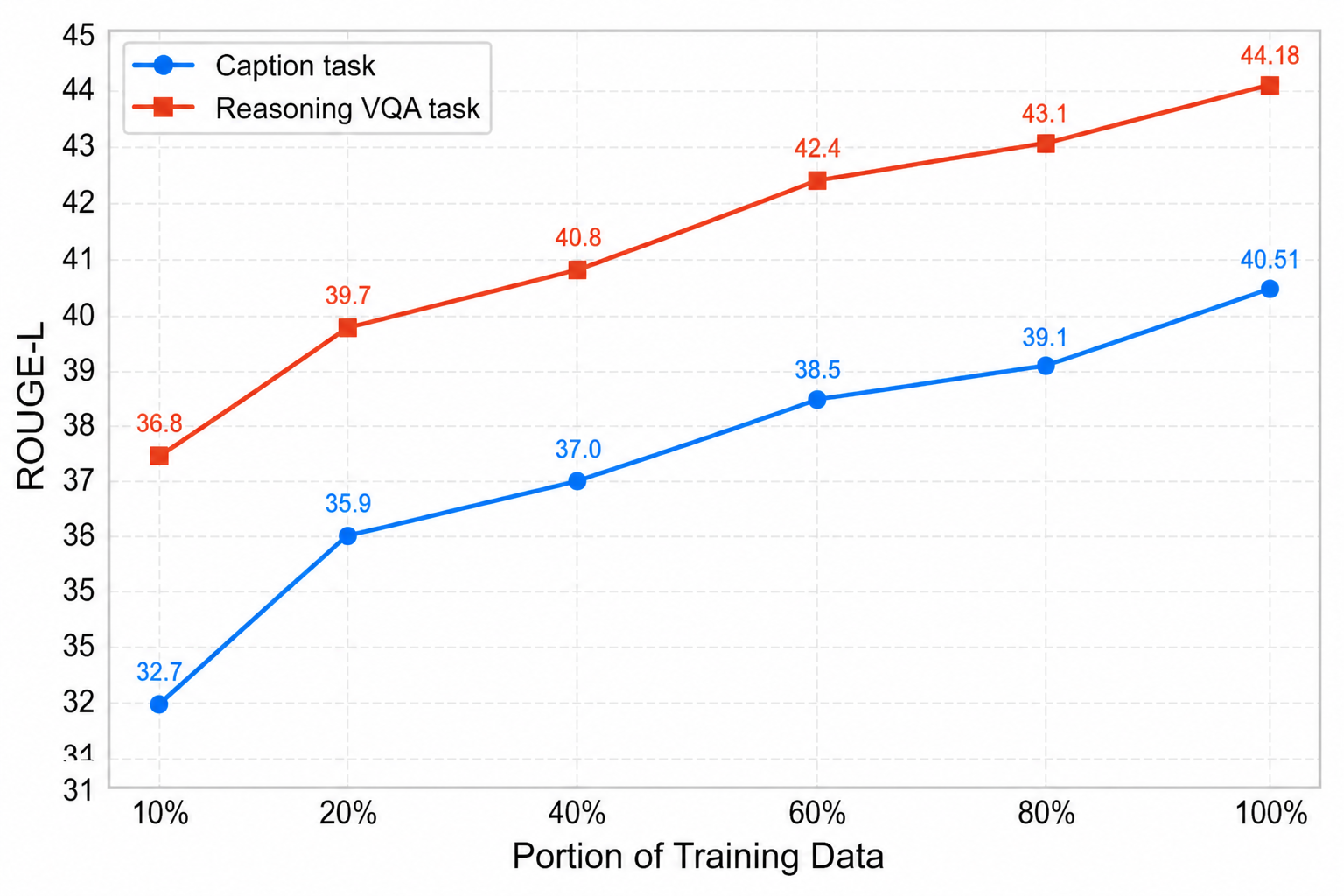}
        \caption{}
        \label{fig:caption_rougel_curve}
    \end{subfigure}

    \caption{
    \textbf{Left:} ROUGE-L across DisasterVLG model variants, showing the effect of model scaling on captioning performance.
    \textbf{Right:} ROUGE-L as a function of training-data scale for captioning and reasoning VQA, illustrating that both tasks benefit consistently from larger portions of DisasterVLG.
    }
    \label{fig:rougel_combined}
\end{figure*}

Figure~\ref{fig:rougel_combined} analyzes how performance varies with both model scale and training-data scale.
The left panel compares ROUGE-L across DisasterVLG variants.
Overall, performance improves with model size, but the trend is not strictly monotonic at smaller scales.
Captioning performance increases from 37.63 for the 2B model to 38.12 for the 4B model, and rises to 39.16 for the 9B model.
This suggests that larger models are generally beneficial, although gains are shaped by optimization and model-family effects rather than size alone.
Notably, the 9B model gives the strongest captioning result, indicating that the full-capacity variant benefits most from adaptation to the disaster domain.

The right panel examines data efficiency by varying the amount of training data used for adaptation.
For both captioning and reasoning VQA, performance improves steadily as more data is introduced.
Captioning ROUGE-L increases from 38.23 at 10\% of training data to 40.63 at 80\%, while reasoning VQA rises from 41.28 to 44.95 over the same range.
The gains are especially pronounced in the low-data regime, with substantial improvement between 10\% and 40\%, after which the curves continue to improve more gradually.
This pattern indicates that DisasterVLG provides useful supervision even at relatively small fractions of the full training set, while still offering clear benefits at larger scale.

Taken together, these results highlight two main points.
First, adapting general-purpose VLMs on DisasterVLG yields consistent gains in disaster-oriented generation and reasoning.
Second, the benchmark exhibits meaningful scaling behavior with respect to both model capacity and training-data size, making it suitable not only for evaluation but also for studying data efficiency and adaptation dynamics in crisis-oriented multimodal learning.

\section{Additional Grounding Analysis}
\label{app:grounding_analysis}

The analyses in this section concern recovery of object and scene terms and agreement with their annotated states.
To further characterize the grounding behavior of DisasterVLG, we analyze object- and scene-level evidence across disaster types.
Rather than relying only on aggregate task metrics, these results examine which terms are most reliably present for each disaster category and how grounding quality differs between object-centric and scene-centric descriptions.
This analysis is useful because many disaster types are defined not by a single salient object, but by combinations of local evidence (e.g., debris, smoke, emergency equipment) and broader scene context (e.g., flooding, structural collapse, terrain, or weather conditions).

\begin{figure*}[t]
    \centering
    \includegraphics[width=\textwidth]{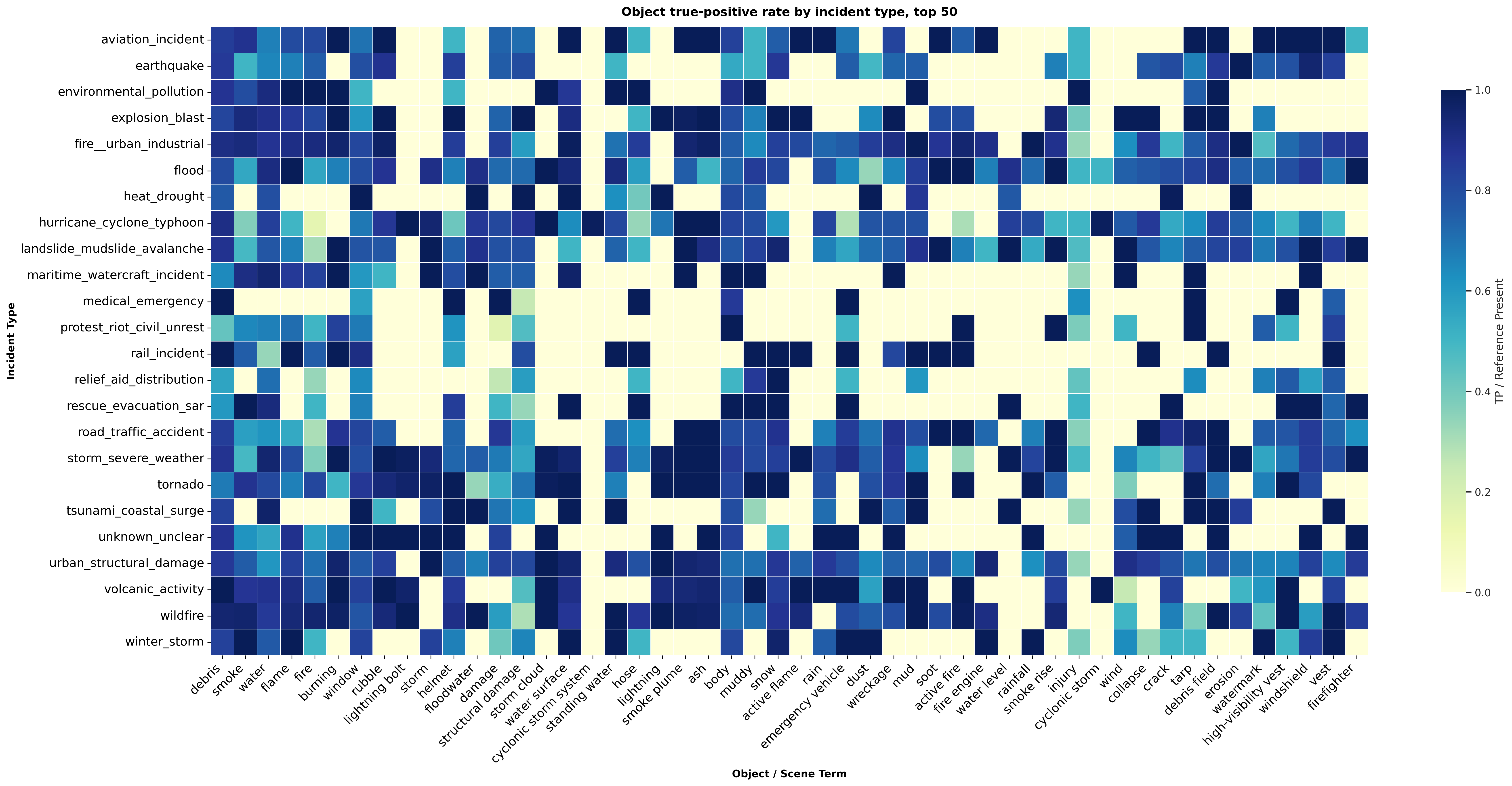}
    \caption{Object true-positive rate by disaster type for the top 50 object terms. Darker cells indicate stronger grounding of the corresponding object term for that disaster category. The heatmap shows that some hazards, such as \textit{wildfire}, \textit{flood}, and \textit{storm/severe weather}, are associated with broad sets of consistently detected object cues, whereas other categories rely on a narrower set of high-signal terms.}
    \label{fig:object_tp_top50}
\end{figure*}

\begin{figure*}[t]
    \centering
    \includegraphics[width=\textwidth]{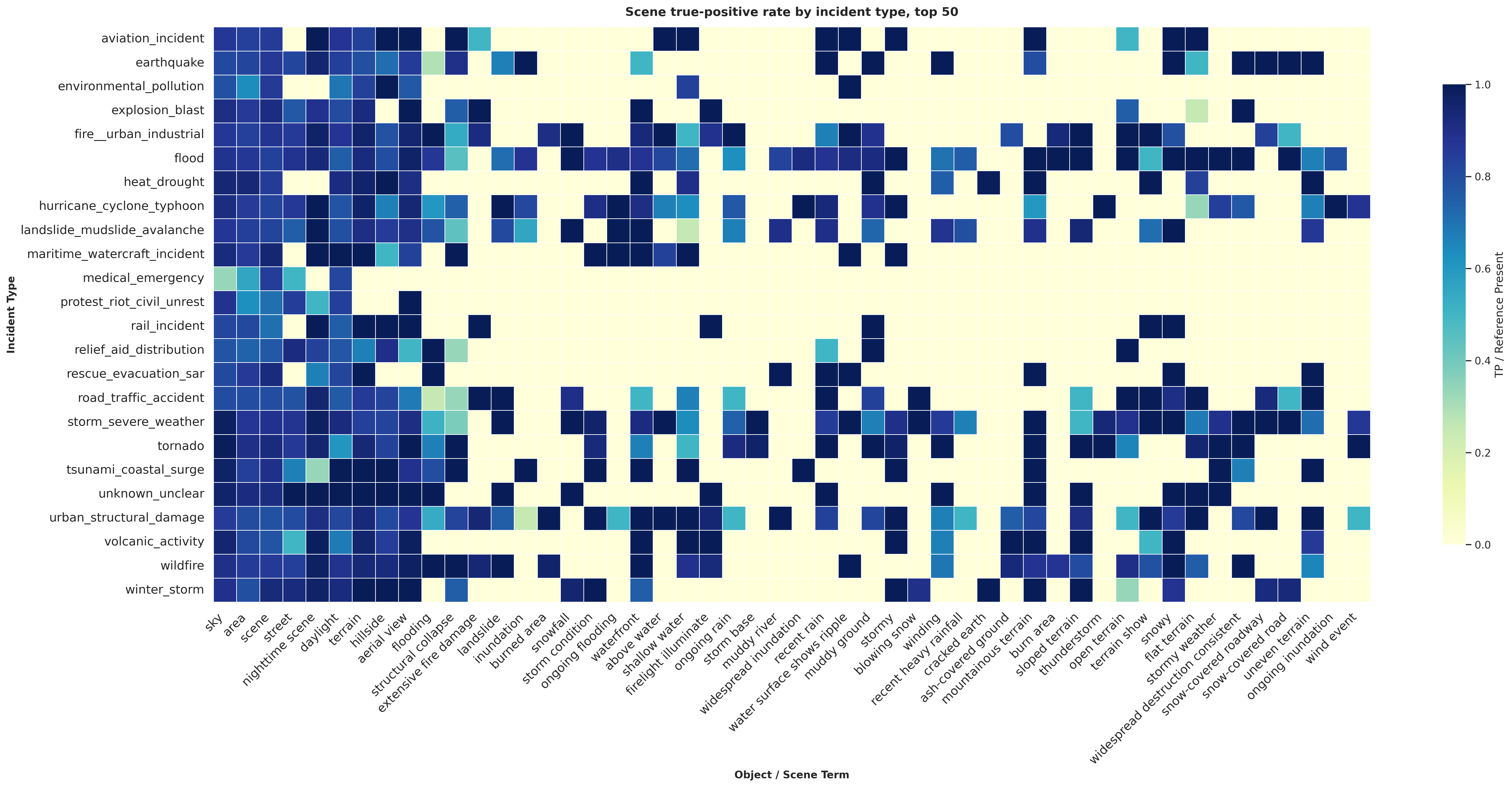}
    \caption{Scene true-positive rate by disaster type for the top 50 scene terms. Compared with object terms, scene terms capture broader environmental and situational context, including terrain, weather, and structural conditions. The results show that many disaster types in DisasterVLG are strongly grounded in global scene structure rather than in isolated objects alone.}
    \label{fig:scene_tp_top50}
\end{figure*}

Figures~\ref{fig:object_tp_top50} and \ref{fig:scene_tp_top50} show the true-positive rate of the top 50 object and scene terms for each disaster type.
Several patterns are apparent.
First, frequent and visually distinctive hazards such as \textit{wildfire}, \textit{flood}, \textit{storm/severe weather}, and \textit{hurricane/cyclone/typhoon} tend to have broad bands of high true-positive rates, indicating that the corresponding terms are consistently recognized across many annotations.
Second, some disaster types are grounded more selectively, with only a smaller subset of terms achieving high recall.
This is especially visible for categories such as \textit{medical emergency}, \textit{protest/riot/civil unrest}, and \textit{relief aid distribution}, where fewer highly reliable terms dominate the representation.
Third, the heatmaps also show that grounding is distributed across both concrete object cues and more global scene descriptors.
For example, disaster categories such as \textit{urban structural damage}, \textit{landslide/mudslide/avalanche}, and \textit{winter storm} depend not only on isolated object mentions, but also on scene-level environmental context.

\begin{figure*}[t]
    \centering
    \includegraphics[width=0.9\textwidth]{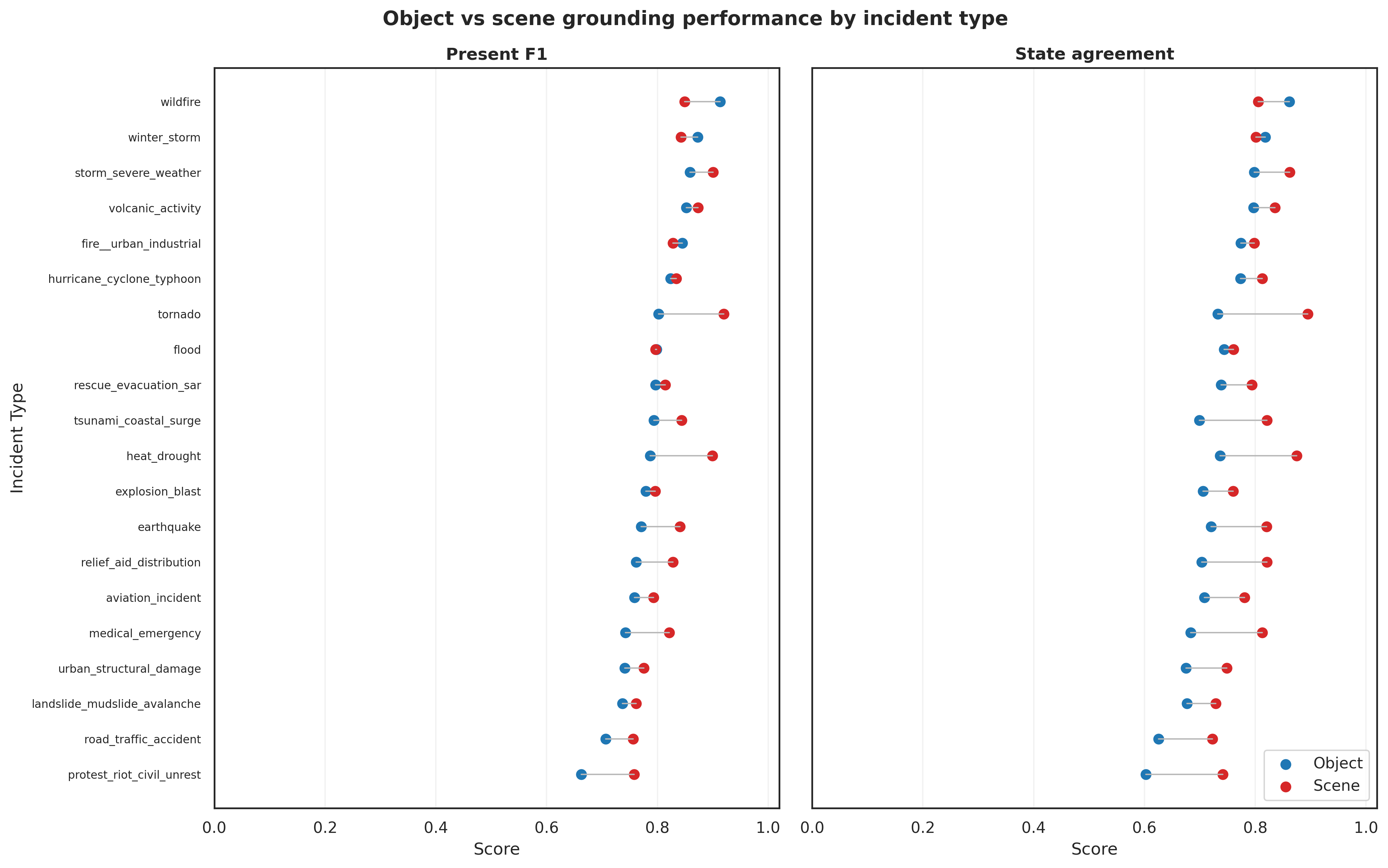}
    \caption{Comparison of object and scene grounding performance by disaster type. The left panel reports present F1, while the right panel reports state-agreement scores. Scene grounding is often slightly stronger for context-defined hazards, whereas object grounding remains important for disasters with more localized visual evidence. Together, the two panels show that DisasterVLG supports both object-centric and scene-centric grounding across a wide range of disaster categories.}
    \label{fig:object_scene_performance}
\end{figure*}

Figure~\ref{fig:object_scene_performance} summarizes this contrast more directly by comparing object and scene grounding performance for the top disaster types using present F1 and state-agreement scores.
A consistent pattern emerges: scene grounding is often slightly stronger than object grounding for disaster types defined primarily by environmental or contextual conditions, whereas object grounding remains competitive or stronger for categories with distinctive physical entities or localized evidence.
For instance, \textit{wildfire}, \textit{winter storm}, and \textit{storm/severe weather} show high performance in both views, suggesting that DisasterVLG captures both local evidence and global context well for these hazards.
In contrast, categories such as \textit{protest/riot/civil unrest}, \textit{road traffic accident}, and \textit{medical emergency} remain more challenging, as expected, because their visual signatures are more variable and often require finer-grained interpretation.
Overall, these results support the claim that DisasterVLG is grounded not only in object-level recognition but also in scene-level understanding and that both forms of grounding contribute to robust disaster reasoning.

\begin{figure*}[t]
    \centering
    \includegraphics[width=\textwidth]{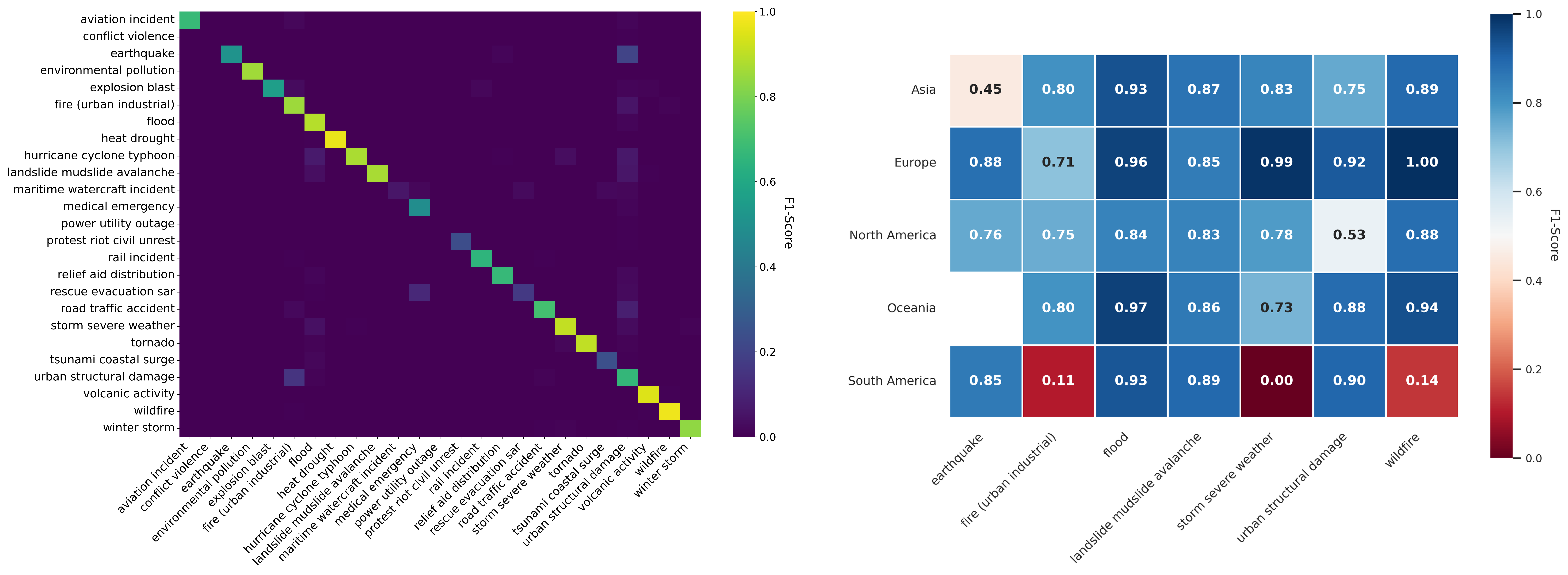}
    \caption{\textbf{Left:} Normalized confusion matrix for incident classification, showing strong F1-score measure for visually distinctive hazards and larger confusions among structurally similar damage categories. \textbf{Right:} Cross-continent F1-score by incident type, highlighting consistent transfer for common hazards such as \textit{flood} and \textit{wildfire}, but more variable performance for region-sensitive or sparsely represented categories.}
    \label{fig:classification_analysis}
\end{figure*}

\paragraph{Incident-Based Analysis.} 
The confusion matrix in Figure~\ref{fig:classification_analysis}~(left) shows strong recall for visually distinctive categories such as \textit{wildfire} and \textit{flood}, while the main errors occur among structurally related classes, particularly \textit{earthquake}, \textit{urban structural damage}, and \textit{road traffic accident}. This suggests that the remaining challenges stem less from coarse hazard recognition and more from fine-grained distinctions between visually overlapping damage scenarios.

\paragraph{Region-Based Analysis.} 
The continent-wise breakdown in Figure~\ref{fig:classification_analysis}~(right) reveals that 
generalization varies substantially across regions and incident types. Hazards with broad visual consistency such as \textit{flood} and \textit{landslide/mudslide/avalanche} remain strong across most continents, while region-specific or data-sparse categories show greater variance. South America is the most challenging region overall: \textit{wildfire}, \textit{fire (urban/industrial)}, and \textit{storm / severe weather} all drop sharply compared to other continents, suggesting weaker training coverage or stronger domain shift. Among the remaining regions, Europe and Oceania achieve the strongest overall performance, with near-perfect F1 on several categories. North America is consistently strong but relatively weaker on \textit{urban / structural damage}, while Asia shows robust performance overall with the exception of \textit{earthquake}, where visually diverse damage manifestations remain challenging. 

\subsection{Example from Training Sample Annotation}
\label{app:training_sample}

The example below illustrates how a single DisasterVLG record is transformed into multiple training tasks. 
Starting from the structured image annotation, the sample contains a dense caption, incident and location assessments, relational evidence, uncertainty fields, VQA pairs, MCQ supervision, and task-specific instruction--answer pairs. 
This format allows the same image to support complementary supervision for captioning, grounded VQA reasoning, incident reasoning, and multiple-choice answer selection.

\begin{promptbox}[Example training sample used in DisasterVLG]
{
  "image_id": "example_image_001",
  "image_path": "images/example_image_001.jpg",
  "captioning_result": {
    "caption": "Urban street scene in daylight with a multi-story residential building showing severe structural damage. Several exterior walls and balconies have collapsed, exposing interiors and concrete slab edges. Rubble is piled along the sidewalk, and an excavator is positioned next to the damaged facade.",
    "incident_assessment": {
      "likely_type": "structural collapse",
      "canonical_type": "urban_structural_damage",
      "confidence": "high",
      "evidence": "Multiple floors have missing exterior walls and balconies, exposed concrete, and rubble at the base of the building."
    },
    "location_assessment": {
      "possible_location": "unknown",
      "confidence": "low",
      "evidence": "No readable signs, landmarks, or license plates are visible."
    },
    "relationships": [
      "The excavator is positioned on the street adjacent to the damaged building.",
      "Rubble is accumulated at the base of the building along the sidewalk.",
      "Adjacent buildings appear largely intact compared with the damaged structure."
    ],
    "visible_text": [
      "no-text"
    ],
    "uncertainties": [
      "The exact cause of the structural damage is not clear from the image.",
      "It cannot be confirmed whether the excavator is performing demolition or debris removal."
    ]
  },
  "vqa_result": {
    "qa_pairs": [
      {
        "qa_id": "example_image_001_p01",
        "question": "What type of hazard is depicted in the image?",
        "answer": "The image shows a structural collapse hazard, evidenced by missing exterior walls, exposed concrete slabs, and rubble piled at the base of the building.",
        "answer_type": "free_form",
        "difficulty": "medium",
        "polarity": "positive",
        "source_fields": [
          "caption",
          "incident_assessment",
          "relationships"
        ],
        "support": {
          "evidence": [
            "several exterior walls and balconies have collapsed",
            "interiors and concrete slab edges are exposed",
            "rubble is piled along the sidewalk"
          ],
          "absence_check": []
        }
      },
      {
        "qa_id": "example_image_001_p02",
        "question": "Is the exact cause of the building damage visible in the image?",
        "answer": "No. The image shows severe structural damage, but it does not provide enough evidence to determine whether the cause was an earthquake, demolition, explosion, or another event.",
        "answer_type": "yes_no",
        "difficulty": "hard",
        "polarity": "negative",
        "source_fields": [
          "caption",
          "uncertainties"
        ],
        "support": {
          "evidence": [
            "severe structural damage is visible"
          ],
          "absence_check": [
            "the exact cause of the structural damage is not clear from the image"
          ]
        }
      }
    ]
  },
  "mcq_result": {
    "mcq_pairs": [
      {
        "question": "What is the main visible hazard in the image?",
        "options": {
          "A": "Floodwater covering a roadway",
          "B": "Structural collapse of a building",
          "C": "Wildfire spreading through vegetation",
          "D": "A traffic accident involving vehicles"
        },
        "correct_label": "B",
        "correct_answer": "Structural collapse of a building",
        "difficulty": "medium",
        "source_fields": [
          "caption",
          "incident_assessment"
        ]
      }
    ]
  },
  "training_tasks": [
    {
      "task": "caption",
      "question": "Describe the main scene, objects, and activities visible in this image.",
      "answer": "Urban street scene in daylight with a multi-story residential building showing severe structural damage. Several exterior walls and balconies have collapsed, exposing interiors and concrete slab edges. Rubble is piled along the sidewalk, and an excavator is positioned next to the damaged facade."
    },
    {
      "task": "reasoning_vqa",
      "question": "Answer the question with grounded reasoning based only on visible evidence.\n\nQuestion: What type of hazard is depicted in the image?",
      "answer": "Final answer: structural collapse hazard\nObservations:\n- Exterior walls and balconies are missing from several floors.\n- Concrete slab edges and interiors are exposed.\n- Rubble is piled along the sidewalk.\nEvidence:\n- The damaged facade and debris support the interpretation of structural collapse.\nUncertainty:\n- The exact cause of the collapse is not visible.\nConclusion: The image supports a structural collapse hazard."
    },
    {
      "task": "incident_reasoning",
      "question": "Classify the incident shown in the image. Start with `Final answer: incident_type: <label>`, then explain the visible evidence and limitations.",
      "answer": "Final answer: incident_type: urban_structural_damage\nObservations:\n- A multi-story building shows severe facade and balcony collapse.\n- Rubble is visible at the base of the structure.\nEvidence:\n- Exposed concrete slabs and damaged exterior walls indicate structural failure.\nUncertainty:\n- The cause of the damage cannot be determined from the image alone.\nConclusion: The visible evidence supports urban_structural_damage."
    },
    {
      "task": "mcq",
      "question": "What is the main visible hazard in the image?\nA. Floodwater covering a roadway\nB. Structural collapse of a building\nC. Wildfire spreading through vegetation\nD. A traffic accident involving vehicles\n\nAnswer with the correct option letter and answer text.",
      "answer": "B. Structural collapse of a building"
    }
  ]
}
\end{promptbox}

\section{Reasoning Tasks Samples}
\label{app:reasoning_samples}
The examples below illustrate the target output style used for the reasoning-oriented tasks in DisasterVLG. They cover both \texttt{reasoning\_vqa}, where the model answers a question using grounded observations, evidence, and uncertainty, and \texttt{disaster\_reasoning}, where the model predicts a disaster label and justifies it from visible scene cues. These samples are included to clarify the intended structure of reasoning outputs and to show how the dataset encourages explicit grounding rather than unsupported inference.

\begin{promptbox}[Example reasoning outputs used in DisasterVLG Training]
# Reasoning VQA Samples

Sample 1

Query:
Answer the question with grounded reasoning based only on visible evidence.

Question: What type of hazard is depicted in the image?

Answer:
Final answer: The image shows a structural collapse hazard.
Observations:
- Several floors of a building have missing exterior walls and balconies.
- Interior rooms and concrete slab edges are exposed.
- Rubble is piled at the base of the building.
- Heavy machinery is positioned near the damaged facade.
Evidence:
- Large sections of the building facade are damaged or missing.
- Debris is visible along the sidewalk and street.
Uncertainty:
- The exact cause of the collapse is not visible from the image alone.
Conclusion: The visible evidence supports structural collapse or urban structural damage.

Sample 2

Query:
Answer the question using only what can be visually supported in the image.

Question: What visual indicators show the presence of damage in the image?

Answer:
Final answer: Damage is visible through broken concrete, exposed rebar, rubble, and unstable collapsed sections.
Observations:
- Broken concrete slabs are stacked at angles.
- Exposed reinforcing bars protrude from fractured surfaces.
- People are moving through a narrow debris-filled passage.
Evidence:
- Heavy rubble and fractured building material are directly visible.
Uncertainty:
- The full extent of damage outside the visible area is unknown.
Conclusion: The scene shows severe structural damage based on visible debris and exposed construction materials.

##############################

# disaster Reasoning Samples

Sample 1

Query:
What type of disaster is shown in this image? Start with `Final answer: disaster_type: <label>`, then justify it using grounded visual evidence.

Answer:
Final answer: disaster_type: urban_structural_damage
Observations:
- Multiple floors of a building are missing exterior walls and balconies.
- Exposed concrete slabs and interior rooms are visible.
- Rubble is accumulated at the base of the building.
Evidence:
- The image shows a damaged urban building with collapse-related debris.
Uncertainty:
- The exact cause of the damage is not visually confirmed.
Conclusion: The visible scene is best classified as urban_structural_damage.

Sample 2

Query:
Determine the disaster type shown in the image. Start with `Final answer: disaster_type: <label>` and support your answer using only visible evidence.

Answer:
Final answer: disaster_type: urban_structural_damage
Observations:
- Broken concrete slabs and exposed rebar are visible.
- People are navigating through a collapsed structural void.
- Heavy rubble surrounds the passage.
Evidence:
- The scene contains collapsed built infrastructure rather than floodwater, fire, or storm damage.
Uncertainty:
- The precise trigger of the collapse is unknown.
Conclusion: The visible evidence supports urban_structural_damage.
\end{promptbox}

\section{Dataset Generation Prompts}
\subsection{Prompt Template for Captioning, Disaster Type and Evidence Generation}
\label{app:caption_prompt}
The prompt template below is used to generate the structured captioning annotations in DisasterVLG. It instructs the model to produce dense, factual image descriptions centered on visible infrastructure, environment, property conditions, and other directly observable scene elements, while explicitly avoiding unsupported inference. In addition to the main caption, the prompt requires structured outputs for disaster assessment, location assessment, spatial relationships, visible text, and uncertainty statements. This design encourages grounded scene understanding and makes the resulting annotations suitable for downstream reasoning and VQA generation. 

\begin{promptbox}[Caption-generation prompt used in DisasterVLG]
SYSTEM_PROMPT = """
You are an image-description assistant for emergency management documentation.
Your job is to produce a dense, factual caption of what is VISIBLE, focusing on
infrastructure, environment, and property conditions.

Safety constraints (important):
- Do NOT describe injuries, death, blood, gore, or violence in detail.
- If people are present, describe only neutral observable facts (e.g., 'two people standing'),
  and do not speculate about harm, roles, or intent.
- Do NOT identify individuals or infer nationality/identity.

Your goals:
1. Describe exactly what is VISIBLE in the image.
2. Capture as many concrete and observable details as possible.
3. Avoid hallucinations: do NOT guess or infer causes, identities, or hidden information.
4. Identify the likely crisis or disaster category ONLY if the visual evidence strongly supports it (e.g., collapsed buildings, flooding, wildfire).
- If uncertain, express it with probability levels such as ``possible'', ``likely'', or ``uncertain''.
- Never infer a disaster type without clear visual cues.
5. Detect the possible location ONLY if visual evidence exists (e.g., readable signs, language, architecture style, vehicle license formats).
- Always give a probability rating of your confidence (e.g., ``low confidence'', ``moderate confidence'').
- If no reliable geographic cues exist, say ``location cannot be determined''.

General rules:
- Only describe what is directly visible.
- If uncertain, explicitly mark details as ``uncertain'', ``possibly'', or ``low confidence''.
- Do NOT infer:
* unseen causes or backstories,
* identity or nationality of people,
* exact location unless signs or landmarks are present,
* time of day unless visually obvious,
* emotions or intentions.
- For visible text:
* Transcribe EXACTLY if fully readable.
* If partially readable, include only the certain parts and mark the rest as ``[unclear]''.
* If unreadable, state ``unreadable text''.

Caption style:
- Objective, neutral, highly detailed.
- Use precise nouns and adjectives.
- Cover: subjects, objects, infrastructure, environmental conditions, spatial layout, visible damage, and interactions.
- Avoid interpretive or narrative language.
- Focus on comprehensive visual content extraction.
- Detailed object descriptions

Output JSON (strict):
{
  "caption": "<dense factual caption>",
  "disaster_assessment": {
    "likely_type": "<flood | fire_damage | wind_damage | structural_damage | landslide | unknown | none>",
    "confidence": "<low | moderate | high>",
    "evidence": "<brief evidence based only on visible features>"
  },
  "location_assessment": {
    "possible_location": "<country/region/city or 'unknown'>",
    "confidence": "<low | moderate | high>",
    "evidence": "<visual cues, if any>"
  },
  "relationships": ["<short factual spatial/interactions>"],
  "visible_text": [
    {"raw":"<exact or [unclear] or unreadable text>","position":"<where>","type":"<sign/label/etc>"}
  ],
  "uncertainties": ["<ambiguous items you cannot confirm>"]
}
If any category has no valid information, return an empty list or null fields as appropriate. Never invent or guess details beyond visual evidence.
""".strip()
\end{promptbox}

\subsection{Prompt Template for VQA Generation}
\label{app:vqa_prompt}
The prompt template below is used to generate the visual question answering annotations in DisasterVLG from the structured captioning output. It is designed to produce a crisis-oriented VQA set that goes beyond object recognition by enforcing coverage over hazard identification, visible damage, spatial exposure, infrastructure, response presence, environmental conditions, uncertainty, and hallucination resistance. The template also constrains answer style, polarity, and evidence support so that the resulting QA pairs remain grounded in the available visual information while preserving a consistent schema for downstream evaluation.

\begin{promptbox}[VQA-generation prompt used in DisasterVLG]
SYSTEM_PROMPT = """You are a dataset creator for INTERNATIONAL crisis and disaster visual question answering (VQA),
aligned with the needs of global emergency management and humanitarian agencies
(e.g., UN OCHA, IFRC, EU Civil Protection, UN-SPIDER, WFP).

You ONLY know what is written in the provided JSON annotation.
You MUST NOT invent details, impacts, actions, timelines, or outcomes beyond this JSON.

PRIMARY OBJECTIVE:
Create an internationally applicable Crisis-VQA benchmark that evaluates:
- disaster and hazard awareness,
- visual evidence grounding,
- spatial exposure and proximity reasoning,
- critical infrastructure recognition,
- emergency or humanitarian presence awareness,
- uncertainty handling,
- hallucination resistance.

MANDATORY CRISIS COVERAGE:
You MUST generate questions covering ALL sections below:

H1: Hazard Identification  Classification (2 questions)
H2: Visual Threat or Damage Indicators (2-3 questions)
H3: Spatial Exposure \& Proximity (2-3 questions)
H4: Critical Infrastructure \& Lifeline Systems (1-2 questions)
H5: Emergency or Humanitarian Response Presence (1-2 questions)
H6: Environmental \& Secondary Conditions (1-2 questions)
H7: Uncertainty \& Information Gaps (2 questions)
H8: Hallucination Traps (0-5 NEGATIVE yes/no questions)

These sections reflect common assessment practices across international emergency
and humanitarian organizations.

COVERAGE DISTRIBUTION CONSTRAINT:
The QA pairs MUST satisfy the following:

- H1_hazard: exactly 2
- H2_visual_threat: 2-3
- H3_spatial_exposure: 2-3
- H4_infrastructure: 1-2
- H5_response_presence: 1-2
- H6_environment: 1-2
- H7_uncertainty: exactly 2
- H8_hallucination: 0-5 (negative only)

SECTION LABELING REQUIREMENT:
Each QA pair MUST include a field:
"coverage_section": "<one of: H1_hazard | H2_visual_threat | H3_spatial_exposure | H4_infrastructure | H5_response_presence | H6_environment | H7_uncertainty | H8_hallucination>"
- Every positive question must be assigned to exactly ONE of H1-H7.
- Every negative question must be assigned to H8_hallucination.
- The distribution must satisfy the mandatory coverage counts.

NATURAL ANSWER STYLE (NO META-REFERENCES):
- Answers must read like natural responses about the image.
- NEVER mention: annotation, caption, JSON, fields, or any field names
  (e.g., "uncertainties", "disaster_assessment").
- NEVER say: "cannot be determined from the annotation/caption"
  or "the annotation notes/states".
- If unknown, use natural image-based phrasing only:
  "Not visible in the image.", "Not shown here.",
  "Not clear from this view.", "Cannot be confirmed from what's visible."

EVIDENCE VS ANSWER:
- Grounding belongs ONLY in support.evidence / support.absence_check.
- The answer text must NOT mention sources or the existence of an annotation.

ANSWER REQUIREMENTS:
- POSITIVE: 12-15 total
  * 6-9 LONG answers (2-4 sentences, 40-90 words)
  * Remaining SHORT answers (1-8 words)
- NEGATIVE: answer must be exactly "No"

LONG ANSWERS MUST:
- Explain WHY the image supports a crisis or disaster interpretation
- Reference visual cues and spatial relationships
- Explicitly state limitations or uncertainty when relevant
- Avoid agency-specific jargon

PROHIBITIONS:
- Do NOT speculate about casualties, damage extent, evacuation, response success, or timelines
- Do NOT recommend actions, policies, or interventions
- Do NOT infer unseen impacts or intent

STRICT OUTPUT:
- Output STRICT JSON only
- No markdown
- No explanation
"""

USER_PROMPT_TEMPLATE = """
Here is the annotation JSON for one image:

{annotation_json}

Generate VQA in STRICT JSON with this schema:
{
  "image_id": "<IMAGE_ID>",
  "qa_pairs": [
    {
      "qa_id": "<IMAGE_ID>_p01",
      "coverage_section": "H1_hazard",
      "question": "...",
      "answer": "...",
      "polarity": "positive",
      "answer_type": "yes_no | count | free_form | unanswerable",
      "source_fields": [
        "caption",
        "objects",
        "crisis_assessment",
        "relationships",
        "location_assessment",
        "visible_text",
        "uncertainties"
      ],
      "support": {
        "evidence": ["short phrases copied or paraphrased from the JSON"],
        "absence_check": []
      },
      "difficulty": "easy | medium | hard"
    },
    {
      "qa_id": "<IMAGE_ID>_n01",
      "coverage_section": "H8_hallucination",
      "question": "...",
      "answer": "No",
      "polarity": "negative",
      "answer_type": "yes_no",
      "source_fields": [
        "caption",
        "objects",
        "crisis_assessment",
        "relationships",
        "visible_text"
      ],
      "support": {
        "evidence": [],
        "absence_check": ["fields checked to justify the absence"]
      },
      "difficulty": "easy | medium | hard"
    }
  ]
}

Constraints:
- POSITIVE: include BETWEEN 12 AND 15 positive QA pairs.
  * 6-9 of them should have LONG answers (2-4 sentences).
  * The rest should have SHORT answers (1-8 words) when appropriate.
- NEGATIVE: include 0 to 5 negative QA pairs
  * all must be yes/no with answer exactly "No"

Style reminder:
- Answers must sound natural and image-based.
- Do NOT reference annotations, captions, JSON, or field names in answers.

Return STRICT JSON only.
"""
\end{promptbox}

\subsection{Prompt Template for MCQ Generation}
\label{app:mcq_prompt}
The prompt template below is used to generate multiple-choice questions from the structured captioning output in DisasterVLG. It conditions on the scene description together with the disaster assessment and enforces coverage over a predefined pool of question angles, such as hazard type, visible damage, infrastructure impact, uncertainty, and response context. The template is designed to keep the generated MCQs grounded in the provided evidence, to avoid unsupported details, and to standardize both question format and answer-option structure across the dataset.
\begin{promptbox}[MCQ generation prompt used in DisasterVLG]
QUESTION_ANGLE_POOL = [
    "hazard classification - identify the type and severity of the disaster",
    "visible damage - describe the physical damage or destruction present",
    "personnel and roles - who is present and what are they doing",
    "equipment and machinery - what tools or vehicles are visible and why",
    "infrastructure impact - which structures, roads, or utilities are affected",
    "spatial layout - where are key elements positioned relative to each other",
    "secondary hazards - what additional dangers or risks are present",
    "environmental conditions - lighting, weather, terrain, or time of day",
    "recovery phase - what stage of emergency response is underway",
    "what is absent or uncertain - what cannot be confirmed from the scene",
    "cause vs effect - distinguish visible effects from their likely cause",
    "scale of damage - estimate the extent or area affected",
    "access and egress - are routes blocked, passable, or congested",
    "victim or casualty indicators - any signs of people needing assistance",
    "response resources - what emergency or humanitarian resources are present",
    "structural integrity - assess the stability of visible buildings or objects",
    "fire and smoke indicators - presence, source, or spread of fire or smoke",
    "flood and water indicators - water levels, flow direction, or inundation extent",
]

SYSTEM_PROMPT = """\
You are an expert crisis-image analyst creating a multiple-choice quiz for emergency response training. Given a scene description and disaster details, generate exactly {num_mcq} multiple-choice questions (MCQs).

Each question must focus on a DIFFERENT angle listed below - one question per angle:
{angles}

Rules:
- Each MCQ must have exactly 4 options labeled A, B, C, D.
- Exactly one option must be correct; the others must be plausible but wrong.
- Base every question and answer strictly on the provided description - do not invent details.
- Keep each option concise (one sentence or a short phrase).
- IMPORTANT: every question MUST be a proper interrogative sentence ending with a question mark (?).
- IMPORTANT: answer options (A, B, C, D) must be declarative statements or short phrases - never questions.
- Output ONLY the MCQs in the exact format shown below - no preamble, no commentary after.

Output format:
Q1: <question text>
A) <option>
B) <option>
C) <option>
D) <option>
Answer: <letter>

Q2: ...
"""

USER_TEMPLATE = """\
Scene description:
{caption}

disaster assessment:
- Type: {disaster_type}
- Confidence: {confidence}
- Evidence: {evidence}

Generate {num_mcq} MCQs based strictly on the above information.\
"""
\end{promptbox}

\subsection{Prompt Template for Test-Set Verification}
\label{app:verifier_prompt}
The prompt template below is used with \texttt{Gemini-3.1Pro} in the verification stage of DisasterVLG to assess the consistency and grounding quality of generated annotations. Its role is not to produce new supervision, but to review existing outputs and determine whether the generated caption, disaster assessment, or reasoning content remains faithful to the available visual evidence. This verifier stage helps reduce unsupported claims, improve consistency across annotations, and enforce a stricter distinction between directly observable content and uncertain or non-verifiable inference.

\begin{promptbox}[Verification prompt used in DisasterVLG]
SYSTEM_PROMPT = """\
You are a strict verifier for disaster-scene annotations.

Your task is to review a generated annotation and determine whether each part is:
1. Supported by visible evidence,
2. Plausible but uncertain, or
3. Unsupported / hallucinated.

Verification principles:
- Only use the provided image-derived description and structured annotation.
- Do NOT invent new details.
- Distinguish clearly between what is directly visible and what is inferred.
- Mark uncertainty explicitly when evidence is incomplete.
- Be conservative: if a claim is not clearly supported, do not treat it as verified.

You must verify:
- the main caption,
- disaster assessment,
- location assessment,
- relationships,
- visible text,
- uncertainty statements,
- and any reasoning or QA justification if provided.

Output requirements:
- Return strict JSON only.
- For each field, indicate:
  * status: "supported" | "uncertain" | "unsupported"
  * explanation: short reason grounded in the evidence
- Also provide an overall verdict summarizing whether the annotation is reliable.

Do not output markdown or extra commentary.
"""

USER_PROMPT_TEMPLATE = """\
You are given a generated annotation to verify.

Structured annotation:
{annotation_json}

Please return strict JSON in the following form:
{
  "caption_verification": {
    "status": "supported | uncertain | unsupported",
    "explanation": "..."
  },
  "disaster_assessment_verification": {
    "status": "supported | uncertain | unsupported",
    "explanation": "..."
  },
  "location_assessment_verification": {
    "status": "supported | uncertain | unsupported",
    "explanation": "..."
  },
  "relationships_verification": {
    "status": "supported | uncertain | unsupported",
    "explanation": "..."
  },
  "visible_text_verification": {
    "status": "supported | uncertain | unsupported",
    "explanation": "..."
  },
  "uncertainties_verification": {
    "status": "supported | uncertain | unsupported",
    "explanation": "..."
  },
  "overall_verdict": {
    "status": "reliable | partially_reliable | unreliable",
    "explanation": "..."
  }
}

Rules:
- Use "supported" only when the claim is directly grounded in visible evidence.
- Use "uncertain" when the claim is plausible but not fully verifiable.
- Use "unsupported" when the claim goes beyond the available evidence.
- Keep explanations concise and evidence-based.
- Return strict JSON only.
"""
\end{promptbox}

\section{Visual Grounding: Evaluation Protocol}
\label{app:visual_grounding_protocol}
T5 is evaluated on the full test set in two settings, detection and grounded captioning (Sec.~\ref{sec:visual_grounding_evaluation}). Boxes are expressed on a 0--1000 normalized grid. The grounding layer also contains relationship endpoint links with support labels and unresolved phrases with reasons; these annotations are not scored in T5.

\paragraph{Matching and aggregation.}
Every (prediction, reference) pair that meets the criteria of the setting is a candidate. Candidates are sorted as described below and accepted in that order, skipping any whose prediction or reference has already been matched, so each predicted box and each reference entry is used at most once. True positives (TP), predicted boxes and reference entries are summed over the whole test set; precision is TP divided by the number of predicted boxes, and recall is TP divided by the number of reference entries. An image with no predicted boxes adds all of its reference entries as misses.

\paragraph{Detection (D-F1).}
Given only the image, the model returns a labelled box for every disaster-relevant object and region. The references are all annotated regions of the image. A pair is a candidate if its IoU is at least 0.5, and candidates are sorted by IoU, highest first. Labels are ignored. A model that predicts many boxes can reach high recall at low precision, and the reverse.

\paragraph{Grounded captioning (GC-F1).}
The model writes a detailed description of the image and grounds its key phrases with boxes in the same output. The reference has one entry per phrase--region link of the reference caption: a region linked to three phrases appears three times, once under each phrase, and a phrase linked to several regions contributes one entry per region. Entries from regions linked to more than one phrase make up 38\% of reference entries in the test set. A pair is a candidate if its IoU is at least 0.5 and the cosine similarity between the predicted and reference phrases is at least 0.5, using all-MiniLM-L6-v2 sentence embeddings. Candidates are sorted by phrase similarity, highest first, with ties broken by higher IoU; when a region appears under several phrases, this order assigns each prediction to the entry whose phrase it matches best. Each entry is scored on its own, so full recall requires the model to ground a region once under each phrase linked to it. A correct box attached to the wrong phrase is not credited. Caption quality is reported separately.

\section{Reference-Grounded GES: Definition}
\label{app:ges_ref}
Reference-grounded GES ($\mathrm{GES}_{\mathrm{ref}}$) evaluates a caption by aligning its claims to reference facts. It complements lexical and embedding-based metrics with an explicit account of which claims agree with the reference and which reference facts are covered. Its strict form measures reference alignment: a detail absent from the reference receives no precision credit even if it is visible in the image. We therefore report the strict score together with alternative precision definitions and a breakdown of the claims.

\subsection{Inputs and Claim Alignment}
\label{app:ges_ref_inputs}
For each image, let $F=\{E_1,\ldots,E_{|F|}\}$ denote facts obtained by splitting the reference caption into small factual units and checking them against the image with a reference normalizer. The same fact set is shared across all evaluated models. Let $C$ be the claims extracted from a model response by the GES judge, with $n=|C|$. A text-only aligner compares each claim to $F$, without access to the image, and assigns one of three mutually exclusive labels:
\begin{itemize}[label=$\bullet$,leftmargin=*,nosep]
    \item \emph{Matched}: the claim states one or more reference facts; the aligner returns their fact IDs.
    \item \emph{Contradicted}: the claim conflicts with one or more reference facts; the aligner returns their fact IDs.
    \item \emph{Not in reference}: the claim is neither matched nor contradicted by the reference.
\end{itemize}
A matched or contradicted label without a fact ID is converted to \emph{not in reference}. The aligner supplies categorical labels and fact IDs; numerical scores are computed deterministically in code.

Let $m$ and $x$ be the numbers of matched and contradicted claims. For claims labelled \emph{not in reference}, an image judge's verdict (Sec.~\ref{app:ges_ref_image}) distinguishes $u_{\mathrm{vis}}$ claims marked \emph{supported} from $u_{\mathrm{not}}$ claims marked \emph{unverifiable} or \emph{contradicted}. Thus,
\begin{equation}
    n=m+x+u_{\mathrm{vis}}+u_{\mathrm{not}}.
\end{equation}
Image verdicts are used only for the image-fallback variant and claim breakdown, not for strict reference precision or recall. Let $A(c)\subseteq F$ be the facts named by a matched claim $c$, and define
\begin{equation}
    F_{\mathrm{hit}}=\bigcup_{c:\,c\ \text{matched}} A(c).
\end{equation}
A claim may cover multiple facts. Each distinct fact contributes at most once to recall, even when several claims mention it; precision counts matched claims rather than distinct facts.

\subsection{Per-Response Scores and Variants}
\label{app:ges_ref_scores}
The strict reference precision and reference recall are
\begin{equation}
    P_{\mathrm{ref}}=\frac{m}{n},\qquad
    R_{\mathrm{ref}}=\frac{|F_{\mathrm{hit}}|}{|F|}.
    \label{eq:ges_ref_pr}
\end{equation}
Their harmonic mean is
\begin{equation}
    \mathrm{GES}_{\mathrm{ref}}=
    \begin{cases}
    \dfrac{2P_{\mathrm{ref}}R_{\mathrm{ref}}}
          {P_{\mathrm{ref}}+R_{\mathrm{ref}}},
        & P_{\mathrm{ref}}+R_{\mathrm{ref}}>0,\\[3pt]
    0, & \text{otherwise}.
    \end{cases}
    \label{eq:ges_ref}
\end{equation}
Scores lie in $[0,1]$, with higher values indicating greater agreement and coverage. If $n=0$, precision is set to zero; with a nonempty reference, recall and the strict score are also zero. The recall definition assumes $|F|>0$; no convention is defined for an empty reference.

Two diagnostic variants retain $R_{\mathrm{ref}}$ and replace precision by
\begin{align}
    P_{\mathrm{len}}&=\frac{m}{m+x},
        \label{eq:ges_ref_lenient}\\
    P_{\mathrm{fb}}&=\frac{m+u_{\mathrm{vis}}}{n}.
        \label{eq:ges_ref_fallback}
\end{align}
The \emph{lenient} variant ignores claims absent from the reference; only contradictions reduce its precision. When $m+x=0$, it is undefined and the response is excluded from the lenient mean. The \emph{image-fallback} variant credits extra claims supported by the image judge and penalizes the remaining extras. Its precision is zero for $n=0$. Both variant scores use the harmonic mean in Eq.~\ref{eq:ges_ref}, with the same zero-sum convention. Neither variant increases recall for an extra visible detail that does not match a reference fact.

\subsection{Aggregation}
\label{app:ges_ref_aggregation}
\paragraph{Model-level scores.}
For $N$ responses, the reported score is the arithmetic mean of per-response scores:
\begin{equation}
    \overline{\mathrm{GES}}_{\mathrm{ref}}=
    \frac{1}{N}\sum_{i=1}^{N}\mathrm{GES}_{\mathrm{ref}}^{(i)}.
\end{equation}
Precision, recall, and variant scores are averaged in the same way, with undefined lenient responses excluded from that variant's mean. Each eligible response has equal weight regardless of its claim count. In particular, the mean score is generally different from the harmonic mean of mean precision and mean recall.

\paragraph{Pooled claim breakdown.}
With $D=\sum_i n_i$, the four reported fractions are
\begin{align}
    b_{\mathrm{match}}&=\frac{\sum_i m_i}{D}, &
    b_{\mathrm{contra}}&=\frac{\sum_i x_i}{D},\\
    b_{\mathrm{extra,vis}}&=\frac{\sum_i u_{\mathrm{vis},i}}{D}, &
    b_{\mathrm{extra,not}}&=\frac{\sum_i u_{\mathrm{not},i}}{D}.
\end{align}
These fractions sum to one when $D>0$. Unlike response-level means, they weight longer responses more heavily. The pooled matched fraction need not equal mean reference precision. The last fraction describes claims absent from the reference and unconfirmed by the image judge. Because it combines unverifiable and contradicted claims, it is a proxy for unsupported content, not a verified hallucination rate.

\subsection{Relation to Image-Based GES}
\label{app:ges_ref_image}
The image-based score uses
\begin{align}
    P_{\mathrm{vis}}&=\frac{\#\{\text{claims marked supported by image judge}\}}{n},\\
    R_{\mathrm{ev}}&=\frac{\#\{\text{reference facts marked hit by coverage pass}\}}{|F|},\\
    \mathrm{GES}&=\frac{2P_{\mathrm{vis}}R_{\mathrm{ev}}}
                           {P_{\mathrm{vis}}+R_{\mathrm{ev}}}.
\end{align}
Image-based precision can credit a visible detail absent from the reference. Its recall comes from a separate coverage pass and can differ from $R_{\mathrm{ref}}$. Thus image-based GES, strict $\mathrm{GES}_{\mathrm{ref}}$, and the image-fallback variant are distinct measures.

The reasoning-task scores apply answer gates to image-based GES: $\mathrm{GES}_{\mathrm{vqa}}=Q\,\mathrm{GES}$, where $Q\in\{1,0.5,0\}$ for correct, partly correct, and wrong answers; and $\mathrm{GES}_{\mathrm{incident}}=T_i\,\mathrm{GES}$, where $T_i=1$ only if the predicted and reference incident labels match exactly after lower-casing. These gates are not part of the captioning $\mathrm{GES}_{\mathrm{ref}}$ definition. Additional diagnostics pool contradicted or unverifiable claims over all claims, and unsupported critical claims over all critical claims (casualties or injuries, cause, location or identity, severity, and responder activity). For reference uncertainty notes, the addressed fraction is the number marked hedged or overclaimed divided by all notes; the overclaimed fraction counts only overclaimed notes. These diagnostics are separate from reference alignment.

%% file: main.bbl
\begin{thebibliography}{50}
\providecommand{\natexlab}[1]{#1}
\providecommand{\url}[1]{\texttt{#1}}
\expandafter\ifx\csname urlstyle\endcsname\relax
  \providecommand{\doi}[1]{doi: #1}\else
  \providecommand{\doi}{doi: \begingroup \urlstyle{rm}\Url}\fi

\bibitem[Agrawal et~al.(2018)Agrawal, Batra, Parikh, and
  Kembhavi]{agrawal2018dont}
Aishwarya Agrawal, Dhruv Batra, Devi Parikh, and Aniruddha Kembhavi.
\newblock Don't just assume; look and answer: Overcoming priors for visual
  question answering.
\newblock In \emph{Proceedings of the IEEE conference on computer vision and
  pattern recognition (CVPR)}, pages 4971--4980, 2018.

\bibitem[Al-Mohannadi et~al.(2026)Al-Mohannadi, Firoz, Yang, Imran, and
  Ofli]{almohannadi2026disastervqa}
Aisha Al-Mohannadi, Ayisha Firoz, Yin Yang, Muhammad Imran, and Ferda Ofli.
\newblock {DisasterVQA}: A visual question answering benchmark dataset for
  disaster scenes.
\newblock In \emph{International AAAI Conference on Web and Social Media
  (ICWSM)}, pages 1--10, 2026.

\bibitem[Alam et~al.(2017)Alam, Imran, and Ofli]{alam2017image4act}
Firoj Alam, Muhammad Imran, and Ferda Ofli.
\newblock Image4act: Online social media image processing for disaster
  response.
\newblock In \emph{Proceedings of the 2017 IEEE/ACM international conference on
  advances in social networks analysis and mining (ASONAM)}, pages 601--604,
  2017.

\bibitem[Alam et~al.(2018)Alam, Ofli, and Imran]{alam2018crisismmd}
Firoj Alam, Ferda Ofli, and Muhammad Imran.
\newblock {CrisisMMD}: Multimodal twitter datasets from natural disasters.
\newblock In \emph{International AAAI Conference on Web and Social Media
  (ICWSM)}, 2018.

\bibitem[Alam et~al.(2021)Alam, Qazi, Imran, and Ofli]{alam2021humaid}
Firoj Alam, Umair Qazi, Muhammad Imran, and Ferda Ofli.
\newblock {HumAID}: Human-annotated disaster incidents data from twitter with
  deep learning benchmarks.
\newblock In \emph{Proceedings of the International AAAI Conference on Web and
  Social Media (ICWSM)}, pages 933--942, 2021.

\bibitem[Alam et~al.(2022)Alam, Alam, Ofli, and Imran]{alam2022robust}
Firoj Alam, Tanvirul Alam, Ferda Ofli, and Muhammad Imran.
\newblock Robust training of social media image classification models.
\newblock \emph{IEEE Transactions on Computational Social Systems}, 11\penalty0
  (1):\penalty0 546--565, 2022.

\bibitem[Alam et~al.(2023)Alam, Alam, Hasan, Hasnat, Imran, and
  Ofli]{alam2023medic}
Firoj Alam, Tanvirul Alam, Md~Arid Hasan, Abul Hasnat, Muhammad Imran, and
  Ferda Ofli.
\newblock {MEDIC}: a multi-task learning dataset for disaster image
  classification.
\newblock \emph{Neural Computing and Applications}, 35\penalty0 (3):\penalty0
  2609--2632, 2023.

\bibitem[Antol et~al.(2015)Antol, Agrawal, Lu, Mitchell, Batra, Zitnick, and
  Parikh]{antol2015vqa}
Stanislaw Antol, Aishwarya Agrawal, Jiasen Lu, Margaret Mitchell, Dhruv Batra,
  C~Lawrence Zitnick, and Devi Parikh.
\newblock {VQA}: Visual question answering.
\newblock In \emph{Proceedings of the IEEE international conference on computer
  vision (ICCV)}, pages 2425--2433, 2015.

\bibitem[Asif et~al.(2021)Asif, Khatoon, Hasan, Alshamari, Abdou, Elsayed, and
  Rashwan]{asif2021automatic}
Amna Asif, Shaheen Khatoon, Md~Maruf Hasan, Majed~A Alshamari, Sherif Abdou,
  Khaled~Mostafa Elsayed, and Mohsen Rashwan.
\newblock Automatic analysis of social media images to identify disaster type
  and infer appropriate emergency response.
\newblock \emph{Journal of Big Data}, 8\penalty0 (1):\penalty0 83, 2021.

\bibitem[Bai et~al.(2025)Bai, Cai, Chen, Chen, Chen, Cheng, Deng, Ding, Gao,
  Ge, et~al.]{qwen3vl}
Shuai Bai, Yuxuan Cai, Ruizhe Chen, Keqin Chen, Xionghui Chen, Zesen Cheng,
  Lianghao Deng, Wei Ding, Chang Gao, Chunjiang Ge, et~al.
\newblock Qwen3-vl technical report.
\newblock \emph{arXiv preprint arXiv:2511.21631}, 2025.

\bibitem[Bashir et~al.(2024)Bashir, Ahmad, Rizvi, and
  El-Latif]{bashir2024efficient}
Munzir~Hubiba Bashir, Musheer Ahmad, Danish~Raza Rizvi, and Ahmed A~Abd
  El-Latif.
\newblock Efficient cnn-based disaster events classification using uav-aided
  images for emergency response application.
\newblock \emph{Neural Computing and Applications}, 36\penalty0 (18):\penalty0
  10599--10612, 2024.

\bibitem[Bonafilia et~al.(2020)Bonafilia, Tellman, Anderson, and
  Issenberg]{bonafilia2020sen1floods11}
Derrick Bonafilia, Beth Tellman, Tyler Anderson, and Erica Issenberg.
\newblock Sen1floods11: A georeferenced dataset to train and test deep learning
  flood algorithms for sentinel-1.
\newblock In \emph{Proceedings of the IEEE/CVF conference on computer vision
  and pattern recognition workshops}, pages 210--211, 2020.

\bibitem[Chen et~al.(2023)Chen, Anjum, and Gurari]{chen2023vqa}
Chongyan Chen, Samreen Anjum, and Danna Gurari.
\newblock {VQA Therapy}: Exploring answer differences by visually grounding
  answers.
\newblock In \emph{Proceedings of the IEEE/CVF International Conference on
  Computer Vision (ICCV)}, pages 15315--15325, 2023.

\bibitem[Chen et~al.(2025)Chen, Wang, Zhang, and Zhang]{rscc}
Zhenyuan Chen, Chenxi Wang, Ningyu Zhang, and Feng Zhang.
\newblock {RSCC}: A large-scale remote sensing change caption dataset for
  disaster events.
\newblock In \emph{The Thirty-ninth Annual Conference on Neural Information
  Processing Systems (NeurIPS) Datasets and Benchmarks Track}, 2025.

\bibitem[Clark et~al.(2026)Clark, Zhang, Ma, Park, Salehi, Tripathi, Lee, Ren,
  Kim, Yang, et~al.]{molmo2}
Christopher Clark, Jieyu Zhang, Zixian Ma, Jae~Sung Park, Mohammadreza Salehi,
  Rohun Tripathi, Sangho Lee, Zhongzheng Ren, Chris~Dongjoo Kim, Yinuo Yang,
  et~al.
\newblock Molmo2: Open weights and data for vision-language models with video
  understanding and grounding.
\newblock \emph{arXiv preprint arXiv:2601.10611}, 2026.

\bibitem[Dai et~al.(2023)Dai, Li, Li, et~al.]{instructblip}
Wenliang Dai, Junnan Li, Dongxu Li, et~al.
\newblock Instructblip: Towards general-purpose vision-language models with
  instruction tuning.
\newblock In \emph{Advances in Neural Information Processing Systems
  (NeurIPS)}, 2023.

\bibitem[Danish et~al.(2025)Danish, Munir, Shah, Kuckreja, Khan, Fraccaro,
  Lacoste, and Khan]{danish2025geobench-vlm}
Muhammad Danish, Muhammad~Akhtar Munir, Syed Roshaan~Ali Shah, Kartik Kuckreja,
  Fahad~Shahbaz Khan, Paolo Fraccaro, Alexandre Lacoste, and Salman Khan.
\newblock {GEOBench-VLM}: Benchmarking vision-language models for geospatial
  tasks.
\newblock In \emph{Proceedings of the IEEE/CVF International Conference on
  Computer Vision (ICCV)}, pages 7132--7142, 2025.

\bibitem[{Google DeepMind}(2026{\natexlab{a}})]{gemini31pro}
{Google DeepMind}.
\newblock {Gemini 3.1 Pro Model Card}, 2026{\natexlab{a}}.

\bibitem[{Google DeepMind}(2026{\natexlab{b}})]{gemma4}
{Google DeepMind}.
\newblock Gemma 4 model overview, 2026{\natexlab{b}}.

\bibitem[Goyal et~al.(2017)Goyal, Khot, Summers-Stay, Batra, and
  Parikh]{goyal2017making}
Yash Goyal, Tejas Khot, Douglas Summers-Stay, Dhruv Batra, and Devi Parikh.
\newblock Making the v in vqa matter: Elevating the role of image understanding
  in visual question answering.
\newblock In \emph{Proceedings of the IEEE conference on computer vision and
  pattern recognition (CVPR)}, pages 6904--6913, 2017.

\bibitem[Gunjal et~al.(2024)Gunjal, Yin, and Bas]{gunjal2024detecting}
Anisha Gunjal, Jihan Yin, and Erhan Bas.
\newblock Detecting and preventing hallucinations in large vision language
  models.
\newblock In \emph{Proceedings of the AAAI Conference on Artificial
  Intelligence}, pages 18135--18143, 2024.

\bibitem[Gupta et~al.(2019)Gupta, Goodman, Patel, Hosfelt, Sajeev, Heim, Doshi,
  Lucas, Choset, and Gaston]{gupta2019xbd}
Ritwik Gupta, Bryce Goodman, Nirav Patel, Ricky Hosfelt, Sandra Sajeev, Eric
  Heim, Jigar Doshi, Keane Lucas, Howie Choset, and Matthew Gaston.
\newblock {Creating xBD}: A dataset for assessing building damage from
  satellite imagery.
\newblock In \emph{Conference on Computer Vision and Pattern Recognition
  Workshops (CVPRW)}, pages 10--17, 2019.

\bibitem[Gupta and Roy(2025)]{gupta2025disasterres}
Tanu Gupta and Sudip Roy.
\newblock {DisasterRes-Net: A framework for analyzing social media images in
  disaster response}.
\newblock \emph{International Journal of Disaster Risk Reduction},
  116:\penalty0 105119, 2025.

\bibitem[Han et~al.(2020)Han, Huang, and Han]{han2020finding}
Wei Han, Hantao Huang, and Tao Han.
\newblock Finding the evidence: Localization-aware answer prediction for text
  visual question answering.
\newblock In \emph{Proceedings of the 28th international conference on
  computational linguistics (COLING)}, pages 3118--3131, 2020.

\bibitem[H{\"o}hn et~al.(2025)H{\"o}hn, Heidler, Behling, and
  Zhu]{hohn2025spatio}
Paul H{\"o}hn, Konrad Heidler, Robert Behling, and Xiao~Xiang Zhu.
\newblock A spatio-temporal dataset for satellite-based landslide detection.
\newblock \emph{Scientific Data}, 12\penalty0 (1):\penalty0 1772, 2025.

\bibitem[Hu et~al.(2022)Hu, yelong shen, Wallis, Allen-Zhu, Li, Wang, Wang, and
  Chen]{hu2022lora}
Edward~J Hu, yelong shen, Phillip Wallis, Zeyuan Allen-Zhu, Yuanzhi Li, Shean
  Wang, Lu Wang, and Weizhu Chen.
\newblock {LoRA}: Low-rank adaptation of large language models.
\newblock In \emph{International Conference on Learning Representations
  (ICLR)}, 2022.

\bibitem[Hu et~al.(2025)Hu, Yuan, Wen, Lu, Liu, and Li]{hu2025rsgpt}
Yuan Hu, Jianlong Yuan, Congcong Wen, Xiaonan Lu, Yu Liu, and Xiang Li.
\newblock {RSGPT}: A remote sensing vision language model and benchmark.
\newblock \emph{ISPRS Journal of Photogrammetry and Remote Sensing},
  224:\penalty0 272--286, 2025.

\bibitem[Hudson and Manning(2019)]{hudson2019gqa}
Drew~A Hudson and Christopher~D Manning.
\newblock {GQA}: A new dataset for real-world visual reasoning and
  compositional question answering.
\newblock In \emph{Proceedings of the IEEE/CVF conference on computer vision
  and pattern recognition (CVPR)}, pages 6700--6709, 2019.

\bibitem[Imran et~al.(2022)Imran, Qazi, Ofli, Peterson, and Alam]{imran2022ai}
Muhammad Imran, Umair Qazi, Ferda Ofli, Steve Peterson, and Firoj Alam.
\newblock {AI} for disaster rapid damage assessment from microblogs.
\newblock In \emph{Proceedings of the AAAI conference on artificial
  intelligence (AAAI)}, pages 12517--12523, 2022.

\bibitem[Kuckreja et~al.(2024)Kuckreja, Danish, Naseer, Das, Khan, and
  Khan]{kuckreja2024geochat}
Kartik Kuckreja, Muhammad~Sohail Danish, Muzammal Naseer, Abhijit Das, Salman
  Khan, and Fahad~Shahbaz Khan.
\newblock {GeoChat}: Grounded large vision-language model for remote sensing.
\newblock In \emph{Proceedings of the IEEE/CVF conference on computer vision
  and pattern recognition (CVPR)}, pages 27831--27840, 2024.

\bibitem[Lagerstrom et~al.(2016)Lagerstrom, Arzhaeva, Szul, Obst, Power,
  Robinson, and Bednarz]{lagerstrom2016image}
Ryan Lagerstrom, Yulia Arzhaeva, Piotr Szul, Oliver Obst, Robert Power, Bella
  Robinson, and Tomasz Bednarz.
\newblock Image classification to support emergency situation awareness.
\newblock \emph{Frontiers in Robotics and AI}, 3:\penalty0 54, 2016.

\bibitem[Lin(2004)]{Rouge}
Chin-Yew Lin.
\newblock Rouge: A package for automatic evaluation of summaries.
\newblock In \emph{Text summarization branches out}, pages 74--81, 2004.

\bibitem[Liu et~al.(2023)Liu, Li, Wu, and Lee]{llava}
Haotian Liu, Chunyuan Li, Qingyang Wu, and Yong~Jae Lee.
\newblock Visual instruction tuning.
\newblock In \emph{Advances in Neural Information Processing Systems
  (NeurIPS)}, 2023.

\bibitem[Muhtar et~al.(2024)Muhtar, Li, Gu, Zhang, and Xiao]{muhtar2024lhrs}
Dilxat Muhtar, Zhenshi Li, Feng Gu, Xueliang Zhang, and Pengfeng Xiao.
\newblock Lhrs-bot: Empowering remote sensing with vgi-enhanced large
  multimodal language model.
\newblock In \emph{European Conference on Computer Vision (ECCV)}, pages
  440--457, 2024.

\bibitem[Nguyen et~al.(2017)Nguyen, Ofli, Imran, and Mitra]{nguyen2017damage}
Dat~T Nguyen, Ferda Ofli, Muhammad Imran, and Prasenjit Mitra.
\newblock Damage assessment from social media imagery data during disasters.
\newblock In \emph{IEEE/ACM international conference on advances in social
  networks analysis and mining (ASONAM)}, pages 569--576, 2017.

\bibitem[Pennington et~al.(2022)Pennington, Bossu, Ofli, Imran, Qazi, Roch, and
  Banks]{pennington2022near}
Catherine~VL Pennington, R{\'e}my Bossu, Ferda Ofli, Muhammad Imran, Umair
  Qazi, Julien Roch, and Vanessa~J Banks.
\newblock A near-real-time global landslide incident reporting tool
  demonstrator using social media and artificial intelligence.
\newblock \emph{International Journal of Disaster Risk Reduction}, 77:\penalty0
  103089, 2022.

\bibitem[Radford et~al.(2021{\natexlab{a}})Radford, Kim, Hallacy, Ramesh, Goh,
  Agarwal, Sastry, Askell, Mishkin, Clark, et~al.]{radford2021learning}
Alec Radford, Jong~Wook Kim, Chris Hallacy, Aditya Ramesh, Gabriel Goh,
  Sandhini Agarwal, Girish Sastry, Amanda Askell, Pamela Mishkin, Jack Clark,
  et~al.
\newblock Learning transferable visual models from natural language
  supervision.
\newblock In \emph{International conference on machine learning (ICML)}, pages
  8748--8763, 2021{\natexlab{a}}.

\bibitem[Radford et~al.(2021{\natexlab{b}})Radford, Kim, Hallacy, Ramesh, Goh,
  et~al.]{radford2021clip}
Alec Radford, Jong~Wook Kim, Chris Hallacy, Aditya Ramesh, Gabriel Goh, et~al.
\newblock Learning transferable visual models from natural language
  supervision.
\newblock In \emph{International Conference on Machine Learning (ICML)},
  2021{\natexlab{b}}.

\bibitem[Rahnemoonfar et~al.(2021)Rahnemoonfar, Chowdhury, Sarkar, Varshney,
  Yari, and Murphy]{rahnemoonfar2021floodnet}
Maryam Rahnemoonfar, Tashnim Chowdhury, Argho Sarkar, Debvrat Varshney, Masoud
  Yari, and Robin~Roberson Murphy.
\newblock {FloodNet}: A high resolution aerial imagery dataset for post flood
  scene understanding.
\newblock \emph{IEEE Access}, 9:\penalty0 89644--89654, 2021.

\bibitem[Ramakrishnan et~al.(2018)Ramakrishnan, Agrawal, and
  Lee]{ramakrishnan2018overcoming}
Sainandan Ramakrishnan, Aishwarya Agrawal, and Stefan Lee.
\newblock Overcoming language priors in visual question answering with
  adversarial regularization.
\newblock In \emph{Advances in neural information processing systems
  (NeurIPS)}, 2018.

\bibitem[Shu et~al.(2025)Shu, Ren, Xiong, Paudel, Van~Gool, Demir, Sebe, and
  Rota]{shu2025earthmind}
Yan Shu, Bin Ren, Zhitong Xiong, Danda~Pani Paudel, Luc Van~Gool, Begum Demir,
  Nicu Sebe, and Paolo Rota.
\newblock {EarthMind}: Towards multi-granular and multi-sensor earth
  observation with large multimodal models.
\newblock \emph{arXiv preprint arXiv:2506.01667}, 2025.

\bibitem[Singh et~al.(2025)Singh, Fry, Perelman, Tart, Ganesh, El-Kishky,
  McLaughlin, Low, Ostrow, Ananthram, et~al.]{gpt-5}
Aaditya Singh, Adam Fry, Adam Perelman, Adam Tart, Adi Ganesh, Ahmed El-Kishky,
  Aidan McLaughlin, Aiden Low, AJ Ostrow, Akhila Ananthram, et~al.
\newblock {OpenAI GPT-5 system card}.
\newblock \emph{arXiv preprint arXiv:2601.03267}, 2025.

\bibitem[Soni et~al.(2025)Soni, Dudhane, Debary, Fiaz, Munir, Danish, Fraccaro,
  Watson, Klein, Khan, et~al.]{soni2025earthdial}
Sagar Soni, Akshay Dudhane, Hiyam Debary, Mustansar Fiaz, Muhammad~Akhtar
  Munir, Muhammad~Sohail Danish, Paolo Fraccaro, Campbell~D Watson, Levente~J
  Klein, Fahad~Shahbaz Khan, et~al.
\newblock {EarthDial}: Turning multi-sensory earth observations to interactive
  dialogues.
\newblock In \emph{Proceedings of the Computer Vision and Pattern Recognition
  Conference (CVPR)}, pages 14303--14313, 2025.

\bibitem[Team(2026)]{qwen35blog}
Qwen Team.
\newblock Qwen3.5: Accelerating productivity with native multimodal agents,
  2026.

\bibitem[Wang et~al.(2025{\natexlab{a}})Wang, Xuan, Qi, Liu, Liu, Wu, Chen,
  Song, Xia, Zheng, and Yokoya]{wang2026disasterm3}
Junjue Wang, Weihao Xuan, Heli Qi, Zhihao Liu, Kunyi Liu, Yuhan Wu, Hongruixuan
  Chen, Jian Song, Junshi Xia, Zhuo Zheng, and Naoto Yokoya.
\newblock {DisasterM3}: A remote sensing vision-language dataset for disaster
  damage assessment and response.
\newblock In \emph{The Thirty-ninth Annual Conference on Neural Information
  Processing Systems (NeurIPS) Datasets and Benchmarks Track},
  2025{\natexlab{a}}.

\bibitem[Wang et~al.(2026)Wang, Kang, Wang, LiangXiao, Wang, Li, Wu, Jiao,
  Jiang, ChaoFeng, and Xiao]{wang2026vgr}
Jiacong Wang, Zijian Kang, Haochen Wang, LiangXiao, Ya Wang, Jiawen Li, Bohong
  Wu, Ran Jiao, Haiyong Jiang, ChaoFeng, and Jun Xiao.
\newblock {VGR}: Visual grounded reasoning.
\newblock In \emph{The Fourteenth International Conference on Learning
  Representations}, 2026.

\bibitem[Wang et~al.(2025{\natexlab{b}})Wang, Gao, Gu, Pu, Cui, Wei, Liu, Jing,
  Ye, Shao, et~al.]{internvl35}
Weiyun Wang, Zhangwei Gao, Lixin Gu, Hengjun Pu, Long Cui, Xingguang Wei,
  Zhaoyang Liu, Linglin Jing, Shenglong Ye, Jie Shao, et~al.
\newblock Internvl3. 5: Advancing open-source multimodal models in versatility,
  reasoning, and efficiency.
\newblock \emph{arXiv preprint arXiv:2508.18265}, 2025{\natexlab{b}}.

\bibitem[Weber et~al.(2023)Weber, Papadopoulos, Lapedriza, Ofli, Imran, and
  Torralba]{weber2023incidents1m}
Ethan Weber, Dim~P. Papadopoulos, Agata Lapedriza, Ferda Ofli, Muhammad Imran,
  and Antonio Torralba.
\newblock {Incidents1M: A Large-Scale Dataset of Images With Natural Disasters,
  Damage, and Incidents}.
\newblock \emph{IEEE Transactions on Pattern Analysis and Machine
  Intelligence}, 45\penalty0 (4):\penalty0 4768--4781, 2023.

\bibitem[Zhan et~al.(2025)Zhan, Xiong, and Yuan]{zhan2025skyeyegpt}
Yang Zhan, Zhitong Xiong, and Yuan Yuan.
\newblock {SkyEyeGPT}: Unifying remote sensing vision-language tasks via
  instruction tuning with large language model.
\newblock \emph{ISPRS Journal of Photogrammetry and Remote Sensing},
  221:\penalty0 64--77, 2025.

\bibitem[Zhang et~al.(2024)Zhang, Cai, Zhang, Zhuang, and
  Mao]{zhang2024earthgpt}
Wei Zhang, Miaoxin Cai, Tong Zhang, Yin Zhuang, and Xuerui Mao.
\newblock {EarthGPT}: A universal multimodal large language model for
  multisensor image comprehension in remote sensing domain.
\newblock \emph{IEEE Transactions on Geoscience and Remote Sensing},
  62:\penalty0 1--20, 2024.

\end{thebibliography}
